# An Analysis of the Value of Information when Exploring Stochastic, Discrete Multi-Armed Bandits

**Isaac J. Sledge[1] and José C. Príncipe[2]**

1  Department of Electrical and Computer Engineering, University of Florida; Computational NeuroEngineering Laboratory (CNEL), University of Florida; isledge@ufl.edu
2  Department of Electrical and Computer Engineering, University of Florida; Department of Biomedical Engineering, University of Florida; Computational NeuroEngineering Laboratory (CNEL), University of Florida; principe@cnel.ufl.edu
*  Correspondence: isledge@ufl.edu, principe@cnel.ufl.edu; Tel.: +x-xxx-xxx-xxxx

The work of the authors was funded via grant N00014-15-1-2103 from the US Office of Naval Research. The first author was additionally funded by a University of Florida Research Fellowship, a Robert C. Pittman Research Fellowship, and an ASEE Naval Research Enterprise Fellowship.

Academic Editor: name
Received: 12 October 2017; Accepted: 16 February 2018; Published: 26 February 2018

**Abstract:** In this paper, we propose an information-theoretic exploration strategy for stochastic, discrete multi-armed bandits that achieves optimal regret. Our strategy is based on the value of information criterion. This criterion measures the trade-off between policy information and obtainable rewards. High amounts of policy information are associated with exploration-dominant searches of the space and yield high rewards. Low amounts of policy information favor the exploitation of existing knowledge. Information, in this criterion, is quantified by a parameter that can be varied during search. We demonstrate that a simulated-annealing-like update of this parameter, with a sufficiently fast cooling schedule, leads to a regret that is logarithmic with respect to the number of arm pulls.

---

## 1. Introduction

There are many environment abstractions encountered in reinforcement learning [1]. One of the simplest examples is the stochastic multi-armed bandit abstraction [2,3]. It is a gambling-based paradigm that can be used to model challenging real-world problems where the environment has a single state and multiple action choices [4].

The stochastic, discrete multi-armed-bandit abstraction consists of a finite set of probability distributions with associated expected values and variances. Each probability distribution is an abstraction of a slot machine, which is colloquially referred to as a one-armed bandit. Initially, each of the distributions is unknown. The agent, a gambler, can choose at discrete time instants to pull one of the slot-machine arms to receive a stochastic reward. The objective of the gambler is to maximize his or her return, which is the sum of the rewards received over a sequence of arm pulls. Since the reward distributions may differ from machine to machine, the objective of the gambler is equivalent to discovering the machine with the best possible pay-off, as early as possible. He or she should then keep gambling using that machine, assuming that the distributions are fixed.

An issue faced in the stochastic multi-armed bandit abstraction is the trade-off between the competing processes of action exploration and exploitation. In some instances, it may be necessary for the gambler to try slot machines that have not been extensively played. More complete knowledge about the pay-off statistics may be obtained, as a result, which can inform the gambler about how to improve the accrued reward. This is referred to as exploration. In other instances, it may be prudent for the gambler to exclusively play on the machine that he or she currently perceives to be the best. This is referred to as exploitation, since the gambler is leveraging his or her existing knowledge about the pay-off statistics to choose an appropriate machine.

A reasonable balance between exploration and exploitation is needed, even for this simple environment. Without an ability to explore, the gambler may fail to discover that one slot machine has a higher average return than the others. Without an ability to exploit, the gambler may fail to pull the best arm often enough to obtain a high pay-off.

Many methods have been developed that attempt to optimally explore for the discrete, stochastic multi-armed-bandit abstraction. One way to quantify their success has been through regret (see appendix A). Regret is the expected total loss in rewards that is incurred due to not pulling the optimal arm at the current episode. Lai and Robbins have proved that a gambler's regret over a finite number of pulls can be bounded below by the logarithm of the number of pulls [5]. There is no gambling strategy with better asymptotic performance.

There are approaches available that can achieve logarithmic regret for stochastic, multi-armed bandits. Prominent examples include the upper-confidence-bound method [6,7] and its many extensions [8,9], Thompson sampling [2,10], and the minimum empirical divergence algorithm [11,12]. The stochastic-based epsilon-greedy [13] and the stochastic-based exponential-weight-explore-or-exploit algorithms [13,14] can also obtain logarithmic regret. Most of these approaches adopt reasonable assumptions about the rewards: they assume that the supplied rewards are independent random variables with stationary means.

In this paper, we propose a stochastic exploration tactic that, for this abstraction, has a weaker assumption on the reward distributions than some bandit algorithms. That is, we assume that the distribution of each new random reward can depend, in an adversarial way, on the previous pulls and observed rewards, provided that the reward mean is fixed. Our approach is based on the notion of the value of information due to Stratonovich [15,16], which implements an information-theoretic optimization criterion.

We have previously applied the value of information to the reinforcement learning problem. In [17], we showed that the value of information could be formulated to explore multi-state, multi-action Markov decision processes. This yielded a Gibbs-based stochastic search strategy. Our simulation results for an arcade-game domain showed that this search strategy outperformed two popular exploration approaches. This was because the value of information aggregates the Markov chains underlying the Markov decision process [18]. It hence performs dynamics reduction during the learning process, which simplifies the search for a high-performing policy. In [19], we extended the value of information to include an uncertainty-based search component so that unexplored regions of the policy space could be explicitly identified and searched. This improved performance was predicated on adjusting the amount of exploration during learning according to a policy cross-entropy heuristic. We showed that this criterion could outperform many uncertainty-based search strategies for arcade-game environments.

In these previous works, we only considered empirical aspects of the value of information. We did not assess its theoretical capabilities. The primary aim and novelty of this paper is to understand how the value of information can implement an optimal-regret search mechanism. We do this for discrete single-state, multi-action Markov decision processes in this paper, so as to make the analysis tractable. This is a crucial first step to showing that this criterion can optimally explore Markov decision processes, as such optimality is not known, given that we are the first to use this criterion for reinforcement learning.

In this setting, the value of information has two equivalent interpretations. The first is that the criterion describes the expected utility gain against the effort to transform the initial arm-selection policy to the optimal selection policy. The second is that it describes the largest increase of average rewards associated with actions that carry a certain amount of information about the environment. In either case, the criterion is defined by two terms that specify a negative free energy in a thermodynamics sense. The first term quantifies the expected rewards of the initially chosen probabilistic policy. The second term measures the divergence between the initial probabilistic policy and the optimal policy. The interplay of these two terms facilitates an optimal search for the best-paying arm under the specified constraints.

Both the policy information and the policy transformation costs are described by a tunable parameter that automatically emerges through the optimization of the value of information criterion. This parameter can be viewed as a kind of inverse temperature. It limits how much the initial policy can transform, which implicitly dictates the amount of exploration. Small values of the parameter near zero impose great constraints on the policy transformation costs. This forces the policy to remain roughly the same as the initial policy. The exploitation of the current best slot-machine lever will therefore dominate. In essence, the amount of information available to the agent about the environment is severely limited, so it is unable to determine if it should change its behaviors. As the parameter value becomes infinite, the expected policy return becomes increasingly important. The transformation costs are essentially ignored for this to occur. The agent becomes free to consider slot-machine arms that differ from the current best policy in a effort to improve regret.

We face certain practical challenges when applying the value of information to multi-armed bandits. One of these entails showing how the criterion can be optimized to produce an action-selection strategy. To deal with this issue, we reformulate the criterion into an equivalent variational form that is convex. We then provide an iterative, coordinate-descent-based approach for uncovering global solutions of this variational expression. The coordinate-descent update leads to a formula for probabilistically weighting action choices that mirrors soft-max-like selection, a common exploration heuristic in reinforcement learning. This weighting process defines a probability that a certain arm will be chosen based upon that arm's expected reward and the value of the inverse-temperature parameter. Arms with higher expected rewards will be chosen more frequently than those with lower average rewards. The final choice of which arm to pull is made stochastically, which spurs an investigation of the action space.

Another challenge that we face is how to achieve optimal regret. We propose to use a simulated-annealing-type schedule for adjusting the inverse-temperature parameter. We prove that the sequence of policy distributions generated under this annealing schedule will converge to a distribution concentrated only on the optimal policies. A sufficiently quick cooling of the inverse-temperature parameter additionally ensures that logarithmic regret can be obtained. This is an interesting finding, as soft-max-like exploration has largely been demonstrated to have either linear-logarithmic or polylogarithmic regret, not logarithmic regret, for the stochastic, multi-armed bandit problem [20,21,24].

The remainder of this paper is organized as follows. We begin, in section 2, with an overview of the multi-armed bandit literature. We predominantly focus on methodologies that rely on stochastic search, as the value of information is also a stochastic approach. We briefly compare the value of information with these methods. Section 3 outlines our methodology. We begin, in section 3.1, by introducing the value of information from a multi-armed bandit perspective. Interpretations for this criterion are provided. In section 3.2, we provide the solution to criterion, which takes the form of a parameterized Gibbs distribution. We modify the Gibbs distribution so as to overcome sampling bias issues that hamper conventional weighted-random styles of exploration. We also provide two potential parameter annealing schedules, each of which has its own advantages and disadvantages. Our theoretical treatment of the value of information is given in appendix A. We prove that two algorithms based on this information-theoretic criterion can yield logarithmic regret for appropriate parameter annealing schedules.

In section 4, we assess the empirical capabilities of the value of information for the bandit problem. We begin this section by covering our experimental protocols in section 4.1. In sections 4.2 and 4.3, we present our simulation results. We assess the performance of the value of information for the fixed-exploration case in section 4.2.1 and the tuned case in section 4.2.2. Performance comparisons against other bandit algorithms are presented in section 4.3. Since the value of information implements a weighted-random search, we consider a variety of stochastic-based sampling techniques, such as $\epsilon$-greedy exploration, soft-max-based selection, and reinforcement comparison. We also compare against the deterministic upper-confidence-bound method. Our results indicate that the value of information outperforms these alternate approaches in a variety of circumstances. Additional comparisons against state-of-the-art techniques are provided in an online appendix. The results show that our information-theoretic criterion is highly competitive and often achieves a similar regret over a finite number of slot-machine plays.

Section 5 summarizes the findings of the paper and reiterates that the novelty of the paper is our analysis of value-of-information criterion. We also outline directions for future research in this section.

## 2. Literature Review

Many exploration schemes have been advanced over the years. Most of these schemes have been developed for rather general models like multi-state, multi-action Markov decision processes. We refer to the survey of Kaelbling et al. [22] for overviews of such schemes.

In this section, we consider popular stochastic exploration tactics for the discrete bandit problem with a discrete number of action possibilities [23]; this is because the value of information implements a stochastic search approach. We note, though, that several other variants of the bandit problem exist. These include adversarial bandits [6,24], non-stationary bandits [6,24], associative bandits [6,25], and budgeted bandits [26], each of which has pertinent exploration-exploitation strategies. Extensions of the bandit problem to the continuous case have also been made [27–29].

One of the most widely employed exploration strategies for discrete, stochastic multi-armed bandits is the $\epsilon$-greedy method [30]. Epsilon-greedy exploration entails choosing either a random arm, with a probability given by the value of $\epsilon$, or the arm with the highest mean reward. For practical problems, the mean reward is estimated from the observed rewards up to the current pull.

The simplest form of $\epsilon$-greedy exploration is the $\epsilon$-first strategy. This approach consists of initially performing an exploration phase during learning, then switching to a pure exploitation phase. In the exploration phase, the arms are randomly selected and pulled during the first $\epsilon k$ rounds, where $k$ is a pre-specified number of rounds. Sub-optimal arms may or may not be dropped from the search process during this stage. During the remaining $(1 - \epsilon)k$ rounds, the lever with the highest estimated mean from the exploration phase is pulled. This is referred to as the exploitation phase. As Even-Dar et al. have shown [31], a log-linear number of pulls $O((|\mathcal{A}|/\alpha^2)\log(|\mathcal{A}|/\kappa))$ during the exploration phase is an upper bound to find an $\alpha$-optimal arm with probability at least $1 - \kappa$. Here, $|\mathcal{A}|$ represents the number of bandit arms. This result can be viewed as an analysis of the asymptotic behavior of the $\epsilon$-first mechanism from a probabilistically-absolutely-correct framework. Mannor and Tsitsiklis [32] provided complementary results for a lower bound and proved that it is also log-linear.

In certain circumstances, context information may be available, at each round, to help determine which arm should be pulled. $\epsilon$-first exploration can also be used for such contextual bandits [33]. More advanced schemes have been proposed that rely on Gaussian processes [34,35] or exponential-weighting for exploration and exploitation using expert advice (Exp4) [24,36].

In the $\epsilon$-first method, the value of $\epsilon$ remains fixed during the learning process. This is a rather restrictive assumption, since it prevents the decision-making process from getting arbitrarily close to the optimal lever. Researchers have proposed so-called $\epsilon$-changing strategies to deal with this issue. Many such methods entail modifying $\epsilon$ across consecutive arm pulls so that the optimal action choice can be made asymptotically. For certain classes of reward distributions, Cesa-Bianchi and Fischer [20] showed that a squared-log regret of $O(\vartheta\log(|\mathcal{A}|/\beta^2)/\beta^2 + \log(k) + \vartheta\log^2(k)/\beta^2)$ can be obtained for $\epsilon$-changing strategies after $k$ rounds. Their approach was referred to as GreedyMix. Here, $\beta, \vartheta$ are parameters, independent of the number of arm pulls $k$, that describe reward statistics for each of the $|\mathcal{A}|$ arms. In GreedyMix, they assumed that $\epsilon$ would be decreased logarithmically with respect to the number of rounds. Auer, Cesa-Bianchi, and Fischer [13] later showed that a log regret of $O(\log(k/\vartheta) + (1 + \beta^2)\vartheta)$ could be obtained for GreedyMix by choosing good initial values for $\epsilon$.

Another widely utilized exploration mechanism is soft-max selection. Soft-max selection entails pulling arms in a stochastic fashion according to a Gibbs distribution. This distribution specifies probabilities for choosing a given lever based upon its expected reward relative to all other possible rewards. Levers with high expected rewards are assigned greater probability mass than those with lower rewards. A single, user-adjustable parameter $\tau$ is present in the distribution that controls the degree of randomness in the lever-selection process. Values of $\tau$ close to zero lead to a greedy selection strategy: the arm with the highest expected reward is consistently favored over all other arms. Values of $\tau$ close to one yield a uniform selection strategy: no action is favored over others, regardless of the expected rewards. $\tau$ is sometimes referred to as temperature, which is due to the connections between the Gibbs distribution and the theory of thermodynamics.

Soft-max selection can be modified into $\tau$-first and $\tau$-changing strategies. In the former case, $\tau$ is set so that exploration is performed only at the beginning of the learning process. The latter case involves monotonically adjusting the temperature $\tau$ from some sufficiently high value to a low value according to either a linear or logarithmic schedule. A logarithmic cooling case was investigated by Cesa-Bianchi and Fischer [20]. They showed that a log-squared regret of $O(\vartheta \log(|\mathcal{A}|/\beta^2)/\beta^2 + \log(k) + \vartheta \log^2(k)/\beta^2)$ can be obtained after $k$ rounds by their SoftMix algorithm. As before, $\beta, \vartheta$ are parameters that depend on the expected values of the reward distributions for the $|\mathcal{A}|$ arms.

A more complicated variant of the soft-max selection, the exponential-weight algorithm for exploration and exploitation (Exp3), was proposed by Auer, Cesa-Bianchi, and Fischer in [13]. Exp3 entails probabilistically choosing a lever based upon an amended Gibbs distribution. The role of the amended distribution is to divide the expected reward by the probability that the action was chosen. Auer et al. showed that a cumulative regret bound of $O((|\mathcal{A}|k\log(|\mathcal{A}|k/\kappa))^{1/2})$, with probability of at least $1 - \kappa$, was possible for finding the $\alpha$-optimal arm when using Exp3. This bound is optimal, as the authors demonstrated that a matching lower bound could be obtained under general circumstances. MacMahan and Streeter [14] later revised Exp3 and Exp4 to take advantage of expert advice and systematically remove irrelevant arms. They furnished a tighter bound of $O((Sk\log(M))^{1/2})$, where $S \ll |\mathcal{A}|$ is a variable that measures the extent to which the expert recommendations agree and $M$ is the number of experts.

In the next section, we provide a soft-max style of action selection for discrete, multi-armed bandits. Our methods are based on the value-of-information criterion proposed by Stratonovich. When applied to reinforcement learning, the value of information quantifies the expected trade-off between policy information and expected rewards. It yields a search process governed by a Gibbs distribution parameterized according to the amount of policy information $\tau_k$. We show that a modified $\tau_k$-changing learning style yields a regret of $O(\log(k))$; this is best regret bound that can be achieved for a stochastic multi-armed bandit problem [4]. For this approach, we assume that knowledge of the mean slot-machine rewards are available to inform the choice of a hyperparameter. This regret bound is an interesting result, as Gibbs-based approaches with a monotone learning-rate sequence have previously only obtained polylogarithmic regret [20,21]. We additionally propose a method with a straightforward means of selecting the hyperparameter. The second approach leads to a regret of $O(\log(k)^{1+2\theta})$, which can be made arbitrarily close to a logarithmic regret by choosing a small $\theta$. No prior knowledge of the reward statistics are needed, unlike in the first method, to achieve this regret bound. This makes it competitive against other Gibbs-based, distribution-free schemes, such as Exp3 and Exp4 [24], which only obtain linear-logarithmic regret.

## 3. Methodology

Our approach to addressing the multi-arm bandit problems follows from the theory of Stratonovich [15, 16]. Stratonovich proposed a mathematical framework for quantitative decision making under information constraints, which took the form of an optimization criterion known as the value of information. The value of information is based on the premise that the inflow of novel information can be used to increase rewards in decision-making problems. That is, the more environmental details that are made available to the agent, the more informed its decisions will be and hence the better the rewards should be. Here, information will take the form of policy transformation costs.

We have previously shown that the value of information can be applied to multi-state, multi-action decision-making problems that can be solved using reinforcement learning [17,19,37]. Here, we simplify this criterion from the multi-state case to that of the single-state so that it is suitable for addressing the multi-armed bandit problem. This is the simplest abstraction that will allow us to see under what conditions the search can be optimal. Extensions of the theoretical results that we obtain for this abstraction may prove useful in the multi-state, multi-action case.

In this setting, we assume that an initial probabilistic arm-selection policy is provided. Such a policy could assign equal probabilities of choosing each arm, for example. The value of information is an optimization problem that seeks a transformed version of this initial policy that yields the highest rewards. The initial policy is not necessarily allowed to change arbitrarily, though, across each arm-pull round. Rather, the transformation is governed by a bounded information constraint term. The lower the bound, the less that the initial policy entries can be modified during a single arm pull. This behavior can impact the obtainable rewards if the initial policy is poor. Conversely, the higher the bound, the more that the initial policy can be adjusted in a given round and hence the greater the potential reward improvement. The degree of policy transformation must be delicately balanced to obtain the best pay-out.

As we will show, the information bounds are dictated by a single parameter that emerges from converting this single-term constrained optimization problem into a two-term unconstrained problem. This parameter simultaneously dictates how much weighted-random-based exploration is performed. We introduce another parameter that adds a uniform-random exploration component, which is crucial for obtaining optimal regret.

It is well known that a constant amount of exploration typically leads to linear regret [20]. A systematic adjustment of exploration is needed to achieve sub-linear regret. We hence consider a simulated-annealing-type approach for tuning the two parameters across each arm pull. As we noted above, we provide two different annealing cooling schedules that switch between near-perfect exploration and near-perfect exploration in a finite number of pulls. We prove in appendix A that these schedules achieve logarithmic regret. Our definition of regret is given in appendix A. We also state our assumptions on the reward distributions in this appendix.

We have provided a table of our variable notation to track the many terms that we define in both this section and the appendix. This table is included at the end of the paper as a supplemental attachment.

### 3.1. Value of Information

In the context of multi-armed bandits, we assume that an agent's initial behavior is described by a prior distribution $p_0(a_k^i)$. Here, $a_k^i \in \mathcal{A}$ represents the $i$th arm pulled in the $k$th round; a single round, which we also refer to as an iteration, is equivalent to one arm pull. The variable $\mathcal{A}$ is the set of all arms.

The agent transforms its behavior to a posterior-like distribution $p(a_k^i)$, which is referred to as the policy. We will also use $\pi_k^i$ to denote the policy distribution. This transformation occurs in a way that optimally trades off the expected utility against the transformation costs for going from $p_0(a_k^i)$ to $p(a_k^i)$. This trade-off is given by the following extremization problem, which is the value of information for the Shannon information case,

$$\max_{p(a_k^i)} \underbrace{\left( \sum_{k=1,2,\dots} \sum_{a_k^i \in \mathcal{A}} p(a_k^i) X_k^i \right)}_{\sum_{k=1,2,\dots} \mathbb{E}_{a_k^i}[X_k^i]} \text{ such that } \underbrace{\sum_{a_k^i \in \mathcal{A}} p(a_k^i) \log \left( \frac{p(a_k^i)}{p_0(a_k^i)} \right)}_{D_{\mathrm{KL}}(p(a_k^i) \| p_0(a_k^i))} \leq \varphi_{\inf}. \tag{1}$$

This expression can be viewed as a problem of finding a version $p(a_k^i)$ of the initial policy that achieves maximal expected rewards in the case where the initial policy $p_0(a_k^i)$ can only change by a prescribed amount. Here, $\varphi_{\inf} \in \mathcal{R}_+$ is a non-negative value that represents the transformation amount bound. The variable $X_k^i \in \mathcal{R}$ represents the reward for the $i$th arm $a_k^i \in \mathcal{A}$ chosen after $k$ arm pulls; it can also be interpreted as a random variable.

The optimization term in (1) can be described in more detail as follows:

> **Optimization Term: Expected Returns.** The term to be optimized represents the expected returns $\sum_{k=1,2,\dots} \mathbb{E}_{a_k^i}[X_k^i] = \sum_{k=1,2,\dots} p(a_k^i) X_k^i$ associated with an arm-selection strategy $p(a_k^i)$ that provides the best pay-out $X_k^i$ over all arm pulls $k$. The selection strategy $p(a_k^i)$ is initially unknown. However, it is assumed to be related to the specified prior $p_0(a_k^i)$, in some manner, as described by the information constraint bound.

As the number of arm pulls becomes infinite, the agent will have complete knowledge of the environment, assuming that the information constraint bound $D_{\mathrm{KL}}(p(a_k^i) \| p_0(a_k^i)) \leq \varphi_{\inf}$ is equal to the action random variable entropy $\varphi_{\inf} = -\sum_{a^j \in \mathcal{A}} p(a^j) \log(p(a^j))$. This term will eventually produce globally optimal pay-outs. Optimal pay-outs can also be achieved if the information constraint bound is annealed at a sufficiently rapid pace. The agent's arm-selection behavior becomes entirely deterministic in such situations, as only the best arm will be pulled. The policy $p(a_k^i)$ becomes a delta function in the limit. Otherwise, the agent's behavior will be stochastic. That is, the policy $p(a_k^i)$ will be described by a discrete unimodal distribution that is not quite uniform and not quite a delta function.

The constraint term in (1) can be interpreted as follows:

**Constraint Term: Transformation Cost.** The constraint term, $D_{\mathrm{KL}}(p(a_k^i) \| p_0(a_k^i))$, a Kullback-Leibler divergence, quantifies the divergence between the posterior $p(a_k^i)$ and the prior $p_0(a_k^i)$. This term is bounded above by some non-negative value $\varphi_{\mathrm{inf}}$, which implies that the amount of overlap will be artificially limited by the chosen $\varphi_{\mathrm{inf}}$.

The value of $\varphi_{\mathrm{inf}}$ dictates by how much the prior $p_0(a_k^i)$ can change to become the posterior $p(a_k^i)$. If $\varphi_{\mathrm{inf}}$ is zero, then the transformation costs are infinite. The posterior will therefore be equivalent to the prior and no exploration will be performed. The expected pay-out will not be maximized for non-optimal priors. If $\varphi_{\mathrm{inf}}$ is larger than the random variable entropy, then the transformation costs are ignored. The prior is free to change to the optimal-reward policy, since the exploration of arms dominates. For values of $\varphi_{\mathrm{inf}}$ between these two extremes, the agent weighs the expected improvement in rewards against the transformation costs. A mixture of exploration and exploitation occurs.

The constraint can also be viewed as describing the amount of information available to the agent about the environment. That is, for small values of $\varphi_{\mathrm{inf}}$, the agent has little knowledge about the problem domain. It is unable to determine how to best adapt its action-selection strategy. As $\varphi_{\mathrm{inf}}$ increases, more information is made available to the agent, which allows it to refine its behavior. In the limit, the agent has the potential for complete knowledge about the environment, which means the action choice will be reward-optimal.

The value of information, given by (1), therefore describes the best pay-out that can be achieved for a specified policy transformation cost. It has the dual interpretation of quantifying how sensitive the arm-selection process will be to environment knowledge, as communicated through the Shannon information term.

This constrained criterion given in (1) can be converted into the following unconstrained variational expression that defines a free-energy difference, in a thermodynamics sense. The unconstrained problem follows from the theory of Lagrange multipliers,

$$\max_{p(a_k^i)} \underbrace{\left( \sum_{k=1,2,\dots} \sum_{a_k^i \in \mathcal{A}} p(a_k^i) X_k^i - \frac{1}{\tau_k} \sum_{a_k^i \in \mathcal{A}} p(a_k^i) \log\left( \frac{p(a_k^i)}{p_0(a_k^i)} \right) \right)}_{\tau_k^{-1} \log\left( \sum_{a_k^i \in \mathcal{A}} p_0(a_k^i) e^{\tau_k X_k^i} \right)}, \tag{2}$$

for some non-negative $\tau_k$. The parameter $\tau_k^{-1}$ sets the relative importance between the transformation cost and the pay-out maximization. $\tau_k^{-1}$ can be viewed as an inverse temperature that can change as the number of arm pulls increases. Adjusting $\tau_k^{-1}$ across each arm pull modifies the amount of exploration that is performed. We discuss ways this can be done, so as to achieve optimal regret, in the next subsection.

The value of information can be posed as the dual problem of maximizing the informational overlap subject to a cost constraint. Both problems are equivalent. For this dual extremization problem, it is easier to see that it convex, as it relies on the minimization of a Kullback-Leibler divergence

$$-\min_{p(a_k^i)} \underbrace{\sum_{a_k^i \in \mathcal{A}} p(a_k^i) \log\left( \frac{p(a_k^i)}{p_0(a_k^i)} \right)}_{D_{\mathrm{KL}}(p(a_k^i) \| p_0(a_k^i))} \quad \text{such that} \quad \underbrace{\left( \sum_{k=1,2,\dots} \sum_{a_k^i \in \mathcal{A}} p(a_k^i) X_k^i \right)}_{\sum_{k=1,2,\dots} \mathbb{E}_{a_k^i}[X_k^i]} \leq \varphi_{\mathrm{cost}} \tag{3}$$

for $\varphi_{\mathrm{cost}} \in \mathcal{R}_+$. The equivalency of (1) and (3) follows from an application of Stjernvall's dominance theory and some additional arguments. We can state this constrained criterion in an unconstrained variational form as a negative free-energy difference,

$$-\min_{p(a_k^i)} \underbrace{\sum_{a_k^i \in \mathcal{A}} p(a_k^i) \log\left( \frac{p(a_k^i)}{p_0(a_k^i)} \right) - \frac{1}{\tau_k} \left( \sum_{k=1,2,\dots} \sum_{a_k^i \in \mathcal{A}} p(a_k^i) X_k^i \right)}_{-\tau_k^{-1} \log\left( \sum_{a_k^i \in \mathcal{A}} p_0(a_k^i) e^{\tau_k X_k^i} \right)}, \tag{4}$$

subject to some non-negative $\tau_k$.

The optimization term in (3) has been analyzed for general multi-state, multi-action reinforcement learning in [17,19]. Here, it can be viewed as follows:

**Optimization Term: Transformation Amount.** The value of information was previously defined by the greatest increase of pay-outs associated with a transformation-cost-constrained policy. The only term in the criterion now measures the transformation costs of the policy. Due to the minimization of the Kullback-Leibler divergence $-D_{\mathrm{KL}}(p(a_k^i)\|p_0(a_k^i))$, the criterion seeks to find a policy $p(a_k^i)$, with a potentially unlimited conversion cost, that can achieve a specified upper bound on the expected pay-out $\sum_{k=1,2,\dots}\mathbb{E}_{a_k^i}[X_k^i]$ over a number of arm pulls.

Likewise, the constraint term in (3) can be interpreted as follows:

**Constraint Term: Pay-Out Limit.** The constraint bounds a policy's range of the expected returns $\mathbb{E}_{a_k^i}[X_k^i]$ by a value $\varphi_{\mathrm{cost}}$. As $\varphi_{\mathrm{cost}}$ approaches infinity, we seek the policy with the highest pay-out. For this to happen, the policy space must be searched finely. Exploration will therefore dominate. As $\varphi_{\mathrm{cost}}$ becomes zero, the constraint to find the policies with the best returns is relaxed. The policy space is searched coarsely, and exploitation dominates. For some value of $\varphi_{\mathrm{cost}}$ between these two extremes, exploitation becomes more prevalent than exploration. The actual value for this to occur will be application-dependent.

The amount of transformation cost can have a profound impact on the resulting action-selection choices. If the cost is high, which corresponds to a policy of limited complexity, then the expected returns for a given policy may be low. This can imply that an agent may not be capable of executing enough actions to complete its objective. Alternatively, when the transformation cost is low, then the agent may therefore be capable of executing more actions than in the former situation. However, such a policy might not be parsimonious, and the agents may take unnecessary actions that can either negatively or neutrally impact the expected returns. For many problems, though, there is a clear region in which the trade-off between policy complexity and simplicity yields meaningful return improvements.

### 3.2. Value-of-Information Policies

The process of decision-making using the value of information proceeds as follows. At the beginning, the agent finds itself in an environment where it is performing optimally, given the information constraints, according to the prior. The agent then experiences a change in the environment, which leads to a change in the rewards. As a consequence of this environmental change, the previous policy is no longer optimal and a new one needs to be found. This requires the maximization of the negative free-energy difference.

It is known that the exponential-based Gibbs distribution maximizes the negative free-energy difference,

$$\underbrace{p(a_{k+1}^i) = \left(\frac{e^{q_k^i/\tau_k}}{\sum_{a^j \in \mathcal{A}} e^{q_k^j/\tau_k}}\right)}_{\pi_{k+1}^i = \text{ exponential component}}, \quad q_{k+1}^i = \frac{X_k^i}{p(a_k^i)}I_k^i, \tag{5}$$

where $q_{k+1}^i$ is an estimate of the expected reward for the $i$th arm $a^i \in \mathcal{A}$. As before, superscripts on variables indicate the index for the particular arm that was pulled, while subscripts on variables indicate the iteration it was pulled.

The update given in (5) can be viewed as a type of coordinate descent where the step-size is automatically chosen at each iteration. We consider the following modification of the Gibbs distribution in (5), which is a kind of mixture model. This was done to ensure that logarithmic regret could be easily obtained,

$$\underbrace{p(a_{k+1}^i) = (1-\gamma_k)\left(\frac{e^{\sum_{s=1}^k q_s^i/\tau_s}}{\sum_{a^j \in \mathcal{A}} e^{\sum_{s=1}^k q_s^j/\tau_s}}\right) + \frac{\gamma_k}{|\mathcal{A}|}}_{\pi_{k+1}^i = \text{ exponential component + uniform component}}, \quad q_{k+1}^i = \frac{X_k^i}{p(a_k^i)}I_k^i. \tag{6}$$

| **Algorithm 1:** $\tau_k$-changing Exploration: VoIMix |
| --- |
| **Input:** $d$: real number in the unit interval such that $0 < d < \min_j \mu^* - \mu^j$, for all $a^j \in \mathcal{A} \backslash a^*$.
**1** Define the sequences $\gamma_k$ and $\tau_k^{-1}$ for $k = 1, 2, \ldots$ by $$\gamma_k = \min\left(1, \frac{5|\mathcal{A}|}{kd^2}\right) \text{ and either}$$ $$\tau_k^{-1} = \frac{1}{|\mathcal{A}|/\gamma_k + 1} \log\left(1 + \frac{d(|\mathcal{A}|/\gamma_k + 1)}{2|\mathcal{A}|/\gamma_k - d^2}\right) \text{ or}$$ $$\tau_k^{-1} = \frac{1}{1 + 2|\mathcal{A}|/\gamma_k} \log\left(1 + \frac{d + 2d|\mathcal{A}|/\gamma_k}{2|\mathcal{A}|/\gamma_k}\right).$$
**2** Let $q^j = 0$ for $j = 1, \ldots, |\mathcal{A}|$.
**3** **for** each $k = 1, 2, \ldots$ **do**
**4** 　　Pull an arm drawn from the distributions $\pi_k^1, \ldots,$ where $$\pi_{k+1}^i = (1 - \gamma_k) \frac{e^{\sum_{s=1}^k q_s^i / \tau_s}}{\sum_{a^j \in \mathcal{A}} e^{\sum_{s=1}^k q_s^j / \tau_s}} + \frac{\gamma_k}{|\mathcal{A}|}.$$
**5** 　　Let $i$ be the index of the pulled arm and $X_k^i$ the obtained reward. Set $q_k^i = X_k^i I_k^i / \pi_{k+1}^i$, where $I_k^i = 1$ and $I_k^j = 0$ for all $j \neq i$. |

| **Algorithm 2:** $\tau_k$-changing Exploration: AutoVoIMix |
| --- |
| **Input:** $\theta$: real number in the range $(0, 0.5)$.
**1** Define the sequences $\gamma_k$ and $\tau_k^{-1}$ for $k = 1, 2, \ldots$ by $$\gamma_k = \min\left(1, \frac{5|\mathcal{A}|}{k} \log^{2\theta}(k)\right) \text{ and}$$ $$\tau_k^{-1} =$$ $$\frac{1}{|\mathcal{A}|/\gamma_k + 1} \log\left(1 + \frac{\log^{-\theta}(k)(|\mathcal{A}|/\gamma_k + 1)}{2|\mathcal{A}|/\gamma_k}\right).$$
**2** Let $q^j = 0$ for $j = 1, \ldots, |\mathcal{A}|$.
**3** **for** each $k = 1, 2, \ldots$ **do**
**4** 　　Pull an arm drawn from the distributions $\pi_k^1, \ldots,$ where $$\pi_{k+1}^i = (1 - \gamma_k) \frac{e^{\sum_{s=1}^k q_s^i / \tau_s}}{\sum_{a^j \in \mathcal{A}} e^{\sum_{s=1}^k q_s^j / \tau_s}} + \frac{\gamma_k}{|\mathcal{A}|}.$$
**5** 　　Let $i$ be the index of the pulled arm and $X_k^i$ the obtained reward. Set $q_k^i = X_k^i I_k^i / \pi_{k+1}^i$, where $I_k^i = 1$ and $I_k^j = 0$ for all $j \neq i$. |

Here, we have introduced a uniform-random exploration component from [13]. This mixture model is parameterized by the inverse temperature $\tau_k^{-1}$ and a mixing coefficient $\gamma_k$. For this update, $X_s^i$ represents the current estimate of the expected reward for arm $a_s^i$, and $q_s^i$ is a version of it that is transformed by the allocation rule $p(a_{k+1}^i) = \pi_k^i$ and an indicator variable $I_s^i$. Observe that this transformed reward is modified by the parameters $\tau_1, \ldots, \tau_s$, up to round $s = k$, and then aggregated across all previous arm pulls $a_1^i, \ldots, a_s^i$, up to round $s = k$. This differs from previous variants of exponential-weighting algorithms, such as soft-max selection, where only the current value of $\tau_k$ modifies the sum of aggregated rewards for arms $a_1^i, \ldots, a_s^i$, up to $s = k$.

Both user-selectable parameters in (6) have the following functionality:

$\tau_k$**: Inverse Temperature Term.** This parameter can be interpreted as an inverse-temperature-like parameter. As $\tau_k^{-1}$ approaches zero, we arrive at a low-temperature thermodynamic system of particles. Such a collection particles will have low kinetic energy and hence their position will not vary greatly. In the context of learning, this situation corresponds to policies $\pi_k^i, \pi_{k+1}^i, \ldots$ that will not change much due to a low-exploration search. Indeed, the allocation rule will increasingly begin to favor the arm with the highest current reward estimate, with ties broken randomly. All other randomly selectable arms will be progressively ignored. On the other hand, as $\tau_k^{-1}$ goes off to infinity, we have a high-temperature system of particles, which has a high kinetic energy. The arm-selection policy $\pi_{k+1}^i$ can change drastically as a result. In the limit, all arms are pulled independently and uniformly at random. Values of the inverse-temperature parameter between these two extremes implement searches for arms that are a blend of exploration and exploitation.

$\gamma_k$**: Mixing Coefficient Term.** This parameter can be viewed as a mixing coefficient for the model: it switches the preference between the exponential distribution component and the uniform distribution component. For values of the mixing coefficient converging to one, the exponential component is increasingly ignored. The resulting arm-choice probability distributions $\pi_{k+1}^i$ become uniform. The chance of any arm being selected is the same, regardless of the expected pay-out. This corresponds to a pure exploration strategy. As the mixing coefficient becomes zero, the influence of the exponential term consequently rises. The values of $\pi_{k+1}^i$ will hence begin dictate how frequently a given arm will be chosen based upon its energy, as described by the reward pay-out, and the inverse temperature. Depending on the value of the inverse-temperature parameter, $\tau_k^{-1}$, the resulting search will either be exploration-intensive, exploitation-intensive, or somewhere in between.

There are a few reasons why we considered this modification of the Gibbs distribution. The foremost reason relates to the temperature parameter: it is difficult to define a continuous parameter update schedule to go from near-perfect exploration to near-perfect exploitation in a finite number of iterations. A means of effectively switching between these two arm-play strategies, in a finite amount of time, is needed in practical situations. The above mixture model is one such way of achieving this goal without sacrificing optimal regret. In particular, it can be shown that the contribution of the Gibbs-distribution component is to force an exploitation of the best-paying arm much more quickly than what occurs for GreedyMix [13]. The above update therefore spends less time unnecessarily exploring the space, which increases the chance of pulling the arm associated with the best-paying slot machine early during learning.

Another reason we utilize this form of the Gibbs distribution is to avoid a well-known exploration flaw. It can be demonstrated that, for many soft-max-type sampling schemes, the optimal arm may not be pulled enough for finite starting values of the inverse-temperature parameter. Sub-par returns may be frequently witnessed. Our model, in contrast, tends to produce sequences of arm pulls that search the action space well. This is due to the incorporation of the uniform distribution component, which acts as an extra exploration factor. The summation of previous expected rewards, versus just the current reward, also helps. It tends to smooth out the rewards from slot machines with high variance but sub-optimal means, thereby ensuring that they are played less often than better machines.

To take advantage of this model, a principled means of choosing both the inverse-temperature and the mixing coefficient parameters is required. One way to do this is by employing a type of simulated annealing. Under this scheme, empirical convergence is realized by letting both parameters monotonically decrease to zero according to some cooling schedule.

We have introduced two possible cooling schedules, in algorithms 1 and 2, which yield optimal and near-optimal regret for the bandit problem, respectively. These schedules are inspired from GreedyMix, but avoid factors that would lead to log-squared regret. For algorithm 1, named VoIMix, we set $\gamma_k = \Theta(1/k)$ and $\tau_k^{-1} = \Theta(\log(k))$. Two cooling schedules are given for $\tau_k^{-1}$ in algorithm 1, both of which yield equivalent regret. It is important to note that algorithm 1 requires the user to supply a real-valued input parameter $d$, which is used to update both $\gamma_k$ and $\tau_k^{-1}$. Selecting initial values of this parameter can be difficult, in practice. It relies on knowledge about the slot machine average rewards and the global-best expected rewards, both of which are unknown to the gambler a priori. In algorithm 2, called AutoVoIMix, the parameter $d$ from algorithm 1 has been removed and is implicitly calculated by way of another real-valued term $\theta$, which is more straight-forward to specify. For values of $\theta$ near zero, algorithm 2 achieves nearly logarithmic regret. A great many arm pulls are needed for this to occur, though. Setting $\theta$ near 0.5 achieves log-squared regret. The slot-machine arms need to be pulled only a few times before this happens.

The parameter update schedules given in these algorithms can be explained as follows. Initially, there is a long period of pure exploration. The length of this period is dictated by $d$, in VoIMix, and $\theta$, for AutoVoIMix; higher values of either parameter increase the period. It also depends on the number of arms, with larger amounts of them leading to a lengthier period. The mixing coefficient, $\gamma_k$, is set to one during this time. The inverse-temperature term, $\tau_k^{-1}$, is set to some value greater than zero. The exponential mixture component is therefore largely ignored in favor of the uniform component. The arm-selection policy $\pi_{k+1}^i$ will sample all of the arms in a uniform manner during this time. This period of exploration corresponds to the case where the inverse-temperature is infinite, if we had used a purely exponential formulation. As the number of arm pulls increases, the weight of the exponential term on the policy $\pi_{k+1}^i$ becomes greater. $\gamma_k$ begins to decay toward zero at a rate determined by both the number of arms and the user-supplied parameter. This, in turn, causes $\tau_k^{-1}$ to decrease toward zero. Higher-paying arms are chosen more frequently, which is due to the increased importance of the exponential component. Eventually, only the arm with the best pay-out will be chosen.

In appendix A, we show that these hybrid allocation rules did not come about arbitrarily. They were designed to yield logarithmic regret for the discrete, stochastic multi-armed bandit problem.

In particular, we apply a variety of inequalities to systematically bound the expected value of the policy probabilities for sub-optimal arms (see proposition A.1). We eventually obtain that $\mathbb{E}[\pi_k^i] = O(1/k)$ for non-optimal arms $a_k^i$ at round $k$ (see proposition A.2). It follows from this bound that the regret is $O(\log(k))$ after $k$ arm pulls. This is an improvement from the regret bound $O(\log^2(k))$ obtained for soft-max-based selection when employing monotone inverse-temperature sequences [20]. Note that it is necessary to perform this regret analysis, as the results obtained by Auer et al. [13] do not automatically extend to our hybrid uniform- and exponential-component update. In [13], the authors only considered a uniform-random exploration term that was driven by a monotone exploration parameter decrease schedule.

For the distribution-free AutoVoIMix, we again systematically bound the expected value of the policy probabilities for sub-optimal arms to get that $\mathbb{E}[\pi_k^i] = O(\log^{2\theta}(k)/k)$ (see proposition A.4). We obtain a regret of $O(\log(k)^{1+2\theta})$ after $k$ arm pulls. This regret bound becomes increasingly equivalent to that of SoftMix selection when $\theta \to 0.5$ and approaches logarithmic regret as $\theta \to 0$. As before, it is necessary to perform this regret analysis. The weak bounds obtained by Auer et al. [24] for their distribution-free approaches were for the case where the exploration remained fixed across each arm pull. In our case, the hyperparameters are adjusted across a given simulation. As well, the exponential terms in their method and our own are different. In either VoIMix or AutoVoIMix, the exponential terms are a function of the previously accrued rewards, not just the immediately received reward.

In appendix A, we also quantify the upper and lower probability bounds for choosing a sub-optimal slot machine arm (see proposition A.3). Such results are important, as they can be used to assess the quality of the exploration process at any arm pull past a given point during learning. Through these bounds, we are also able to demonstrate that the value of information has a much higher chance of pulling the optimal arm than other schemes, including that of Auer et al. [13]. This finding theoretically illustrates the benefit of combining a uniform-random component with the exponential component that arises naturally through optimizing the value-of-information criterion. Attempting to create a purely Gibbs-distribution-based exploration scheme that yields logarithmic regret is a difficult problem that has yet to be solved [21].

All of the results obtained for VoIMix and AutoVoIMix apply to the adversarial bandit algorithm without any modification [4]. This is due to the general assumption that we make on the reward distributions, which is outlined in appendix A.

### 3.2.1. Automated Hyperparameter Selection

It is difficult, a priori, to arrive at good hyperparameter values $d$ and $\theta$. An option is to set and iteratively adapt the hyperparameters during the learning process, similar to what we did with the other terms.

A heuristic that we have found to work well involves the use of the policy cross-entropy [19],

$$\mathbb{E}_{\pi_k^i}\left[ -\log\left( (1-\gamma_k)\left( \frac{e^{\sum_{s=1}^k \tau_s q_s^i}}{\sum_{a^j \in \mathcal{A}} e^{\sum_{s=1}^k \tau_s q_s^j}} \right) + \frac{\gamma_k}{|\mathcal{A}|} \right) \right] = -\sum_{a^i \in \mathcal{A}} \pi_k^i \log(\pi_{k+1}^i).$$

The policy cross-entropy is a bounded measure of the agent's learning rate: it can be viewed as the rate of change for probabilistic policies across arm pulls $k$ and $k + 1$. If the arm-selection probabilities change by a great amount, then the cross-entropy will be near one. The acquisition of new behaviors is therefore prevalent, which typically precedes steep rises pay-outs across a few episodes. If arm-selection probabilities change only slightly, then the cross-entropy will be near zero. Existing behaviors are hence retained, leading to moderate or negligible improvements in rewards.

Such connections suggest an efficient learning scheme. A low hyperparameter value should be used, in the beginning, to survey the entire action space. Reasonably high-performing arms be quickly uncovered when doing so. Eventually, the cross-entropy will dwindle and reach a near-steady-state value. Before this occurs, the hyperparameter should be gradually increased to promote a more thorough investigation of the policy space. This determination can be done using threshold tests. If no further improvement is possible over a reasonable period, then the hyperparameter should be steadily decreased toward its minimum valid value. We will use this scheme in some of our simulations.

This scheme is conceptually supported by statistical thermodynamics. It is known that the entropy of an isolated system can only increase. This leads to the conclusion that entropy is maximized at thermal equilibrium. When the system is not isolated, the maximum entropy principle is replaced by the minimum free energy principle. In the context of the value of information, the Lagrangian is the free energy. This principle states that the Lagrangian cannot increase and therefore reaches its minimum value at thermal equilibrium. By annealing the hyperparameter, the system is always kept either at or near equilibrium. At zero temperature, the system reaches the minimum energy configuration. This corresponds to the situation where we have the best policy.

## 4. Simulations

In the previous section, we provided an information-theoretic criterion for quantifying the expected increase in rewards associated with transformation-constrained policies. We also developed a means for optimizing this criterion, which provides a probabilistic mechanism for choosing a given arm based on its past history of expected rewards.

In this section, we quantify the empirical performance of the criterion. The aims of our simulations are multi-fold. First, we want to assess the relative performance of our value-of-information-based exploration approach with existing ones. We would also like to gauge how well our results align with theoretical expectations. Secondly, we want to determine how the value-of-information search mechanism investigates the problem domain.

Toward this end, we consider a series of difficult bandit problems. We demonstrate that the value-of-information-based VoIMix and AutoVoIMix can effectively address these problems. We first consider the case where the two exploration parameters of VoIMix and AutoVoIMix are fixed to constant values across the rounds. Four inverse-temperature parameter values are used to establish the baseline performance and illustrate the effects of a certain amount of weighted-random exploration on each problem. Three mixing coefficient values are used to understand how uniform-random search can aid in finding the best-paying arm by avoiding the sampling bias issue. We then compare these results to the case where the parameters are allowed to adapt according to the cooling schedules given in the previous section. This is done to highlight the regret improvement and empirically justify the two algorithms.

Both VoIMix and AutoVoIMix rely on a hyperparameter that defines the length of a pure-exploration search period. From a theoretical standpoint, the value of this hyperparameter is immaterial, as all valid values will yield the same regret in the limit. Empirically, the hyperparameter value can impact the short-term pay-outs, as we show. This is an important caveat for investigators looking to apply the value of information to practical problems. Choosing good hyperparameter values can be difficult, though, even for AutoVoIMix. We therefore rely on a cross-entropy heuristic for tuning the values across each arm pull. We show that adjusting the hyperparameters yields lower regret and higher pay-outs than the fixed-value case. This indicates that either of our methods could be well suited for real-world bandit applications.

We consider five other bandit algorithms for comparative purposes. Two of these algorithms, $\epsilon$-greedy and soft-max, were chosen due not only to their popularity, but also their commonalities with the value of information. Stochastic, soft-max-based derivatives, such as pursuit methods and reinforcement comparison, are also employed. Both approaches rely on different search mechanisms, which offers experimental variety. Lastly, we consider the deterministic, tuned and un-tuned upper confidence bound approaches. Our simulation results reveal that the value of information outperforms these methods either in general or in specific instances. Explanations for this behavior are provided so that practitioners can determine when to potentially employ a given methodology.

### 4.1. Simulation Preliminaries

The difficulty of the multi-armed bandit problem is fully characterized by two attributes: the distribution properties used to model the slot machines and the number of slot-machine arms.

For our simulations, the rewards for each of the slot-machine arms were sampled from normal distributions. We originally considered a range of additional distributions, such as uniform, inverse normal, and triangular, but found that the results were highly similar. We therefore only report the findings for the normal distribution.

The means of the slot-machine reward distributions were chosen uniformly from the unit interval and fixed for the duration of the simulation. We evaluate the algorithms for a reward distribution variance of $\sigma^2 = 1$, which is fixed. The value corresponds to standard deviations that are one-hundred percent of the interval containing the expected values. This yields a complex problem no matter the number of arms, as the reward distributions have a wide spread and overlap significantly. The agent must therefore judiciously sample each of the arms so that the one with the best average pay-out can be identified and consistently pulled.

Another aspect that changes the difficulty of the problem is the number of slot machines. We evaluate the algorithms for three, ten, and thirty slot machine arms. Three arms leads to fairly easy tasks, while ten arms furnishes marginally difficult ones. Over thirty arms begins to provide tasks that are challenging. Originally, we considered greater numbers of slot machine arms. However, we found that their relative behaviors are mostly consistent with with the thirty-armed case.

For the value of information, we assume that an initial policy is supplied. The results in the appendix indicate that regret is independent of the initial policy. Nevertheless, we specify that the initial policies have uniform probabilities, so as to not introduce bias that could lead to simpler problems.

### 4.2. Value of Information Results and Analysis

#### 4.2.1. Fixed-Parameter Case

We assess the performance of the value of information in two situations: when the parameters $\tau_k$ and $\epsilon_k$ remain fixed across arm pulls during learning and when they are updated automatically. Short-term simulation results for the former case are presented in figure 4.1. Note that for all of the regret plots in this section, we divide the regret, at each round, by the number of arm pulls. Therefore, this modified regret should decrease toward zero as the number of rounds increases.

**Exploration Amount and Performance.** Figure 4.1(a) contains probability simplex plots for VoIMix. Such plots highlight the arm-selection probability distributions of the iterates during the learning process for the three-armed bandit problem. They indicate how VoIMix is investigating the action space across all of the Monte Carlo trials and if it is focusing on sub-optimal arms.

The optimal action for these three-armed bandit problems is associated with the bottom, left corner of the simplex. We hence expect to see a purple-colored region near this corner of the plot, which would indicate that the optimal action is selected frequently during learning. All of the other corners should have either red- or orange-colored regions, as these actions should not be consistently chosen after a series of initial arm pulls.

The value of information operates under the premise that more information can lead to an increase in accrued rewards due to a more thorough exploration of the action space. The plots given in figure 4.1(a) show that the empirical results are aligned with these expectations, at least for the three-armed bandit case. For a fixed low inverse-temperature $\tau_k^{-1}$, VoIMix is reserved in its search due to the effects of the information constraint on the prior. The iterates briefly explore the two sub-optimal arms on the top and right sides of the simplex before congregating on the left half of the simplex. The distribution for the left-hand plot of figure 4.1(a) indicates that the iterates quickly converge to the optimal arm. For high $\tau_k^{-1}$ values, larger portions of the simplex are investigated before converging to the optimal arm. Between $\tau_k^{-1} \approx 0.15$ and $\tau_k^{-1} \approx 1.0$, the simplex distribution does not change greatly for this problem.

Figure 4.1 also contains plots of the average regret as a function of time, the average reward, and the cumulative chance that the optimal arm is chosen. The first two criteria summarize how the methods handled the problem. The third criterion is relevant to assess for applications in which minimizing the number of sub-optimal arm plays is crucial. The averages reported in these plots were derived from a thousand Monte Carlo simulations so as to eliminate the bias of any overly poor or good runs. They are also a function of different parameter choices that correspond to varying amounts of exploration.

Figures 4.1(b) and (c) highlight the effects of the inverse-temperature parameter $\tau_k^{-1}$ on the performance. As $\tau_k^{-1}$ decreases, the episode-averaged regret tends to drop rapidly during the first few arm pulls. Likewise, the average reward increases more rapidly. The effect becomes much more pronounced as the number of bandit arms is small. Both occurrences are a product of quickly finding either the best-paying arm or well-paying arms. As the number of arms increases, there is a reversal in this trend: larger values of $\tau_k^{-1}$ begin to yield better long-term results due to the increased exploration that is needed to sample all of the arms. We present additional results in tables B.1(a) and (b) of the online appendix.

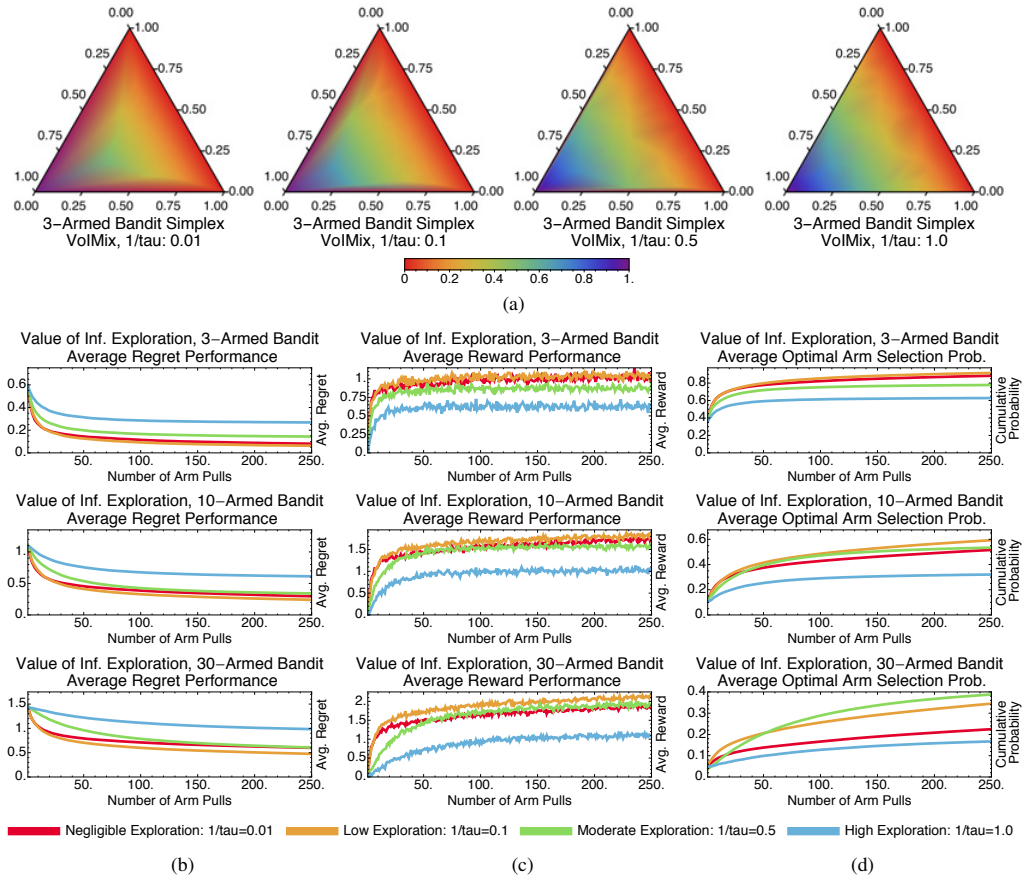

(a)

(b)                                    (c)                                    (d)

**Figure 4.1:** Results for the value-of-information-based VoIMix for the fixed-parameter case. The plots in (a) highlight the regions in the action probability simplex that are visited during the simulations for the three-armed bandit problem. Cooler colors (purple, blue and green) correspond to probability triples that are encountered often during learning. Cooler Warmer (yellow, orange, and red) correspond to probability triples that are not seen often. The correct action is the one associated with the bottom, left corner of the simplex. The simplex plots were produced for fixed parameter values and averaged across independent simulations. The plots in (b), (c), and (d) give the regret, reward, and optimal-arm selection cumulative probability for the three-, ten-, and thirty-armed bandit problem when the distribution variance is $\sigma^2 = 1$. The red, orange, green, and blue curves correspond to fixed inverse temperatures of $\tau^{-1} = 0.01$, 0.1, 0.5, and 1.0, respectively.

Figure 4.1(d) captures the cumulative probability of choosing the optimal arm. These plots show that the fixed-parameter case of VoIMix was unable to reliably find the best-paying arm with a high probability. As the number of arms increased, the short-term cumulative probability dropped, regardless of the amount of exploration $\tau_k^{-1}$ that was used. It can be seen, however, that higher amounts of exploration more frequently uncover the optimal arm. This occurrence aligns with the preceding regret and reward results.

The inverse-temperature parameter $\tau_k^{-1}$ is not the only way to control the search amount in VoIMix. The mixing coefficient $\gamma_k$ also dictates how finely the action space will be investigated. For the simulations in figure 4.1, we assumed that the mixing coefficient was 0.1. One in ten arm pulls would therefore be random. Setting it higher, for these problems, often led to increasingly more exploration than was necessary. The rewards dropped and the regret was inflated, the results of which are presented in table B.1(c). Setting the mixing coefficient lower also impacted the results, as the pay-out became worse. Such results are given in table B.1(d).

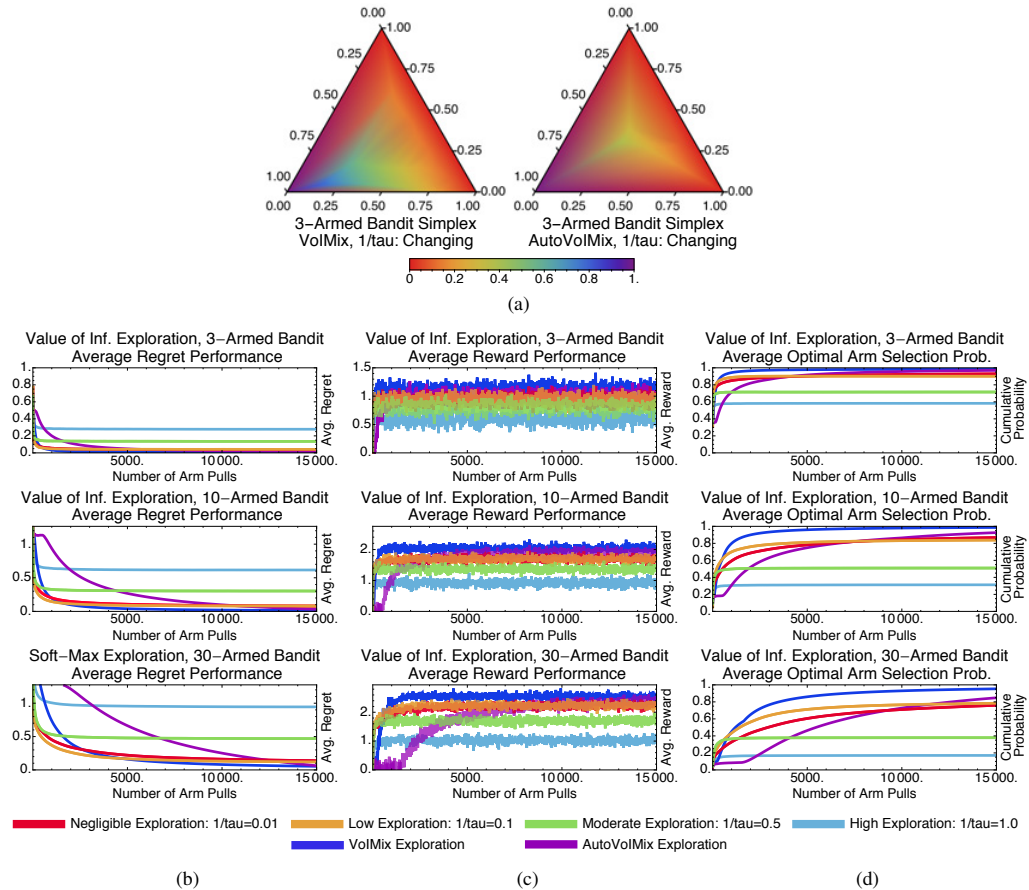

Figure 4.2: Long-term results for the value-of-information-based VoIMix and AutoVoIMix for the fixed and variable parameter case. The hyperparameters $d$ and $\theta$ are fixed for each method. The plots in (a) highlight the regions in the action probability simplex that are visited during the simulations for the three-armed bandit problem. The plots in (b), (c), and (d) give the regret, reward, and optimal-arm selection cumulative probability for the three-, ten-, and thirty-armed bandit problem when the distribution variance is $\sigma^2 = 1$. In these plots, the purple curves correspond to VoIMix where the inverse-temperature and mixing parameters are allowed to vary throughout the simulation, according to the cooling schedule given in the previous section. The dark blue curves correspond to AutoVoIMix where the inverse-temperature and mixing coefficient are allowed to vary throughout the simulation, according to the cooling schedule given in the previous section. The red, orange, green, and blue curves correspond to fixed inverse temperatures of $\tau^{-1} = 0.01$, 0.1, 0.5, and 1.0, respectively.

### 4.2.2. Adaptive-Parameter Case

The results given above are for the case where the parameter values were constant throughout the simulation. This is useful for understanding the algorithmic behaviors in different scenarios. Leaving the parameters fixed can, however, lead to linear regret in the worst case. It is therefore not indicative of the true capabilities of VoIMix, which is known to have logarithmic regret.

We now consider VoIMix and AutoVoIMix when the temperature changes across each arm pull. We also highlight the effects of the hyperparameters and show the performance improvements that can be obtained when choosing them automatically.

**Exploration Amount and Performance.** In figure 4.2, we provide results for both VoIMix and AutoVoIMix over a longer number of iterations. These results are for the cases where $\tau_k^{-1}$ and $\gamma_k$ are either fixed or automatically tuned according to the cooling schedules given in the previous section.

Plots of the probability simplexes for the tuned versions of VoIMix and AutoVoIMix are given in figure 4.2(a). As before, these plots highlight the distribution of iterates during the learning process. The iterates should converge to the bottom, left corner of the simplex, which is the optimal action for the three-armed bandit case. This occurs for VoIMix: after about 250 to 750 arm pulls, there is high chance selecting the optimal arm. The blue- and green-colored region in the bottom, left corner of the simplex plot corroborates this claim. The dark-purple-colored region in the left, bottom corner of the simplex indicates that the probability of selecting the optimal arm becomes near-unity and stays that way throughout a majority of the learning process. The yellow- and orange-colored portions of the plot indicate that VoIMix heavily considered the arm associated with the bottom-right corner of the simplex early during learning. This was the second-best paying arm in all of our Monte Carlo trials.

Different results were observed for AutoVoIMix. The corresponding probability simplex plot in figure 4.2(a) captures that a large portion of the iterates cluster near the middle of the simplex. Each arm has a near-equiprobable chance of being chosen. This region of yellow and orange colors arises from the iterates for the first 5000 arm pulls, which is when a majority of the exploration is performed. Eventually, the optimal arm was identified. The iterates traced an almost direct path from the middle of the simplex to the bottom, left corner. This path has a slightly more green color than the paths to the other two corners, suggesting that arm probabilities from this part of the simplex were consistently selected later in the learning process. The probability for choosing the optimal arm was high near the end of learning, which led to the purple-colored region in this corner.

As shown in figure 4.2(b), the regret for the adaptive-parameter case typically improved beyond the results of the fixed-parameter case after several hundred to thousand iterations. The rate at which the regret dropped depended on the number of slot-machine arms. Such events signaled that a high-paying arm was reliably uncovered and consistently chosen with a high probability, which can be gleaned from figure 4.2(d). The total rewards also jumped, after a few thousand arm pulls, compared to the best results for the fixed-parameter case. This can be seen in figure 4.2(c). This event coincided with the end of the perfect exploration period and a decay of the inverse-temperature parameter and mixing coefficient. Quantitative comparisons showing that the variable-parameter case yielded better performance than the fixed case are provided in table B.2(a) in the appendix.

A drawback of VoIMix is that setting the single hyperparameter can prove difficult. This is because it relies on knowledge of the reward distributions to which gamblers are not privy. We introduced AutoVoIMix as a way to circumvent this issue. However, AutoVoIMix tends to choose subpar arms compared to the non-fixed-parameter version of VoIMix, which can be seen in figure 4.2(d). On average, the regret and total rewards are worse, compared to VoIMix, when relying on parameter annealing, as indicated in figures 4.2(b) and 4.2(c). Quantitative comparisons between the fixed-parameter VoIMix case and variable-parameter AutoVoIMix case are provided in table B.2(b) in the appendix.

**Hyperparameter Values and Performance.** Both VoIMix and AutoVoIMix have a single hyperparameter that must be chosen by the investigator. For VoIMix, we set hyperparameter value $d_k$ to halfway between zero and the smallest mean reward difference. For AutoVoIMix, we also set the hyperparameter $\theta_k$ in the middle of the valid range. This offered a reasonable period for exploration and hence yielded good performance.

Reducing $d_k$ or increasing $\theta_k$ typically shortened the exploration phase, which led to one of two outcomes. One possibility was that the optimal arm was sampled frequently enough for it to be consistently selected after the main exploration phase. The regret was reduced compared to the results shown here. This occurred for bandits with a small number of arms. The other possibility, which occurred frequently, was that an unnecessary amount of arm exploration was performed, especially when dealing with only a few slot-machine arms. The optimal arm could be reliably identified in this case. However, the regret was penalized due to the agent not finding and sticking with the optimal arm sooner. This is highlighted in tables B.3(a) and B.4(a) in the appendix.

Raising $d_k$ in VoIMix or reducing $\theta_k$ in AutoVoIMix had a similar effect as this latter case. It led to lengthier periods of exploration. Results are provided in tables B.3(b) and B.4(b) in the appendix. We observed that VoIMix consistently outperformed AutoVoIMix, regardless of the hyperparameter values.

In the previous section, we introduced a cross-entropy-based scheme for adapting the hyperparameters. When using this scheme, the total regret and accrued pay-out for VoIMix improved compared to the situation where the best-performing hyperparameter value was fixed a priori. This is illustrated in table B.5(a) in the appendix. For AutoVoIMix, we witnessed a similar behavior; results are provided in table B.5(b) in the appendix. In either case, we increased the hyperparameter value by twenty percent, over a window of thirty episodes, every time the cross-entropy fell below one third. Once the cross-entropy changed by no more than ten percent across fifty episodes, we decreased the hyperparameter value by thirty percent every twenty episodes.

### 4.2.3. Fixed- and Variable-Parameter Case Discussions

Our simulations have demonstrated that VoIMix and AutoVoIMix are capable of effectively addressing the discrete multi-armed bandit problem. We now discuss aspects of these methods' results.

**Exploration Amount Discussions.** We considered two possible instances of VoIMix: one where its parameters were fixed another where they varied. Our simulations showed that the latter outperformed the former. This was because a constant amount of exploration might explore either too greatly or not thoroughly enough to achieve good performance. Without the optimal annealing schedule that we provided, substantial trial and error testing would be needed to find reasonable exploration amounts for practical problems.

Our simulation results highlight that the automatically tuned VoIMix outperforms the automatically tuned AutoVoIMix for these bandit problems. This was also witnessed when adjusting the methods' hyperparameters according to the cross-entropy heuristic. Such results align with our theoretical analyses. AutoVoIMix can only approach logarithmic regret when its hyperparameter value $\theta_k$ tends toward zero. VoIMix can achieve logarithmic regret in the limit, no matter the value of its hyperparameter $d_k$ in the valid range.

We have additionally provided distribution plots of the arm-selection probabilities overlaid on the simplex. These plots highlighted that the best-paying arm can be uncovered when the temperature $\tau_k^{-1}$ is either fixed or automatically adjusted. We only considered simplex distribution plots for the three-armed bandit problem. It is not possible to produce exact plots of the probability simplex for bandit problems with more arms, since those probability vectors often cannot be losslessly embedded in a three-dimensional domain.

When manually lowering the temperature, the iterates became more focused and targeted the optimal arm more frequently. This is because the temperature value ensures that only the best-paying arm is chosen with a high probability. When raising the temperature, the search process near-uniformly considered a larger portion of the simplex. The chance of choosing an arm approaches that of any other arm. Even when considering $\tau_k^{-1} = 1.0$, which represents a high amount of policy change, the optimal arm could still be uncovered about half of the time. The corresponding simplex plot shows that the iterates do routinely pull the optimal arm after a few thousand plays.

For our simulations, we sorted the arm rewards so that the best-playing arm is in the bottom-left corner of the simplex and the worst playing arm is in the top-middle corner. We thus expected that the iterates would tend to cluster on the bottom half of the simplex. However, the worst-paying arm appears to be played just as frequently as the second-best paying arm. Additional analysis revealed that this was due to the high variance of the reward distributions: it was often possible for the worst arm to have a better rewards over many arm pulls than the second-best arm. Both arms would therefore have a high probability of being chosen across multiple iterations. The choice of one of these arms over the other changed frequently, which is why the search process near-uniformly investigated these regions of the simplex values of $\tau_k^{-1}$ away from one.

When automatically tuning the inverse temperature according to VoIMix, the resulting simplex plot indicates that it quickly starts to play the optimal arm. Only the two best-paying arms are routinely sampled, which did not happen in the fixed-parameter case. This materialized due to a steep drop in the exploration rate during training, which would emphasize the selection of arms with high expected rewards. We witnessed this behavior for problems with many more arms. The tuned version of VoIMix will thus likely perform well for real-world bandit problems. It automatically modifies the amount of search exploration in proportion to the intrinsic complexity of the problem. This is a major advantage compared to manually choosing the exploration amount.

For the tuned AutoVoIMix, the center regions of the simplex are more thoroughly explored before the optimal arm is consistently pulled. Such investigations may yield little useful insights about the slot-machine distributions after a few hundred to thousand arm pulls. Many arm pulls will hence be required before the regret can be decreased. This can prove problematic for application domains where the physical time and effort needed to supply rewards to the agent is high; the utility of drug-based treatments in clinical trials is one such example.

Both VoIMix and AutoVoIMix can achieve the same long-term rewards and appear to reliably identify the optimal arm. However, VoIMix does so at a much faster rate, leading to a better short-term regret. If practitioners can rely on auxiliary knowledge about their bandit problems to choose good hyperparameter values, then VoIMix should be favored compared to AutoVoIMix. Otherwise, AutoVoIMix should be utilized, with the understanding that the ease of choosing its hyperparameter is traded off against a, potentially, poor initial regret.

**Hyperparameter Value Discussions.** An issue faced when using many multi-armed bandit algorithms, including VoIMix and AutoVoIMix, is the proper setting of hyperparameter values by investigators. Such hyperparameters often have a great impact on the corresponding results. They are also difficult to set so that the corresponding regret bounds hold.

To help avoid choosing poor values, we provided a semi-automated scheme of adapting the hyperparameters $d_k$ and $\theta_k$ based upon the policy cross-entropy This led to improved empirical performance. In the case of VoIMix, this improvement occurred because the hyperparameter $d_k$ was often close to the upper bound of the average reward difference $\min_j \mu^* - \mu_j$. Learning happens very quickly over the first few arm pulls compared to when the values of $d_k$ are close to zero. For AutoVoIMix, $\theta_k$ was set to $\theta_k \approx 0.5$, which likewise led to a quick and effective survey of the space. The value of $\theta_k$ was then iteratively decreased toward zero to promote a more judicious sampling of arms that can lead to near-logarithmic regret in the limit.

The reason why this hyperparameter adjustment approach worked is because the rate of reward improvement is connected with the policy cross-entropy. The rate of cost reduction depends on both the current inverse temperature and mixing coefficient values and hence the hyperparameter value. Any changes made to the hyperparameter cause a deformation of the system's free energy from the Shannon information term in the Lagrangian, when the inverse temperature is zero, to the distortion term. It is therefore obtaining the global minimizer of the value of information at low values of the inverse temperature and tracking the evolution of this minimum as the inverse temperature parameter is increased by way of the hyperparameter.

Lower hyperparameter values are considered to be better for AutoVoIMix. As the hyperparameter decreases toward zero, so does the power term of the logarithm that defines both the mixing coefficient and the inverse temperature. This causes the regret to eventually approach optimality. A great many arm pulls are needed for this to occur, though, unlike in VoIMix. Raising the hyperparameter value reduces the number of pulls needed for the regret to become optimal. The regret becomes log-squared, however, which leads to worse behaviors than for the baseline case. The expected theoretical performance should be no better than soft-max selection in this situation. Empirically, AutoVoIMix manages to accrue a higher total pay-out than the soft-max-based SoftMix in such instances, as we will show in the next subsection. This could be attributed to the fact that the AutoVoIMix regret has a smaller-magnitude constant factor. The increase in performance could also be due to the additional exploration factor in the probability mass assignment.

### 4.3. Methodological Comparisons

To provide context for these results, we have compared the value of information to other commonly used approaches for addressing the bandit problem. These include epsilon-greedy, soft-max, pursuit, and reinforcement-comparison exploration, all of which are stochastic. We also consider the deterministic upper-confidence-bound method. As before, the results given are averaged over a thousand Monte Carlo trials.

We also provide comparisons of the value of information against state-of-the-art techniques including the minimum empirical divergence algorithm [11,12], Thompson sampling [2,10], and various extensions of the upper-confidence-bound method [8,9]. Our results indicate that the value of information is competitive against these techniques, as it often supplies similar regret. It also pulls sub-optimal arms as infrequently as these alternate schemes. Due to page constraints, the corresponding simulations and discussions are provided in an online appendix.

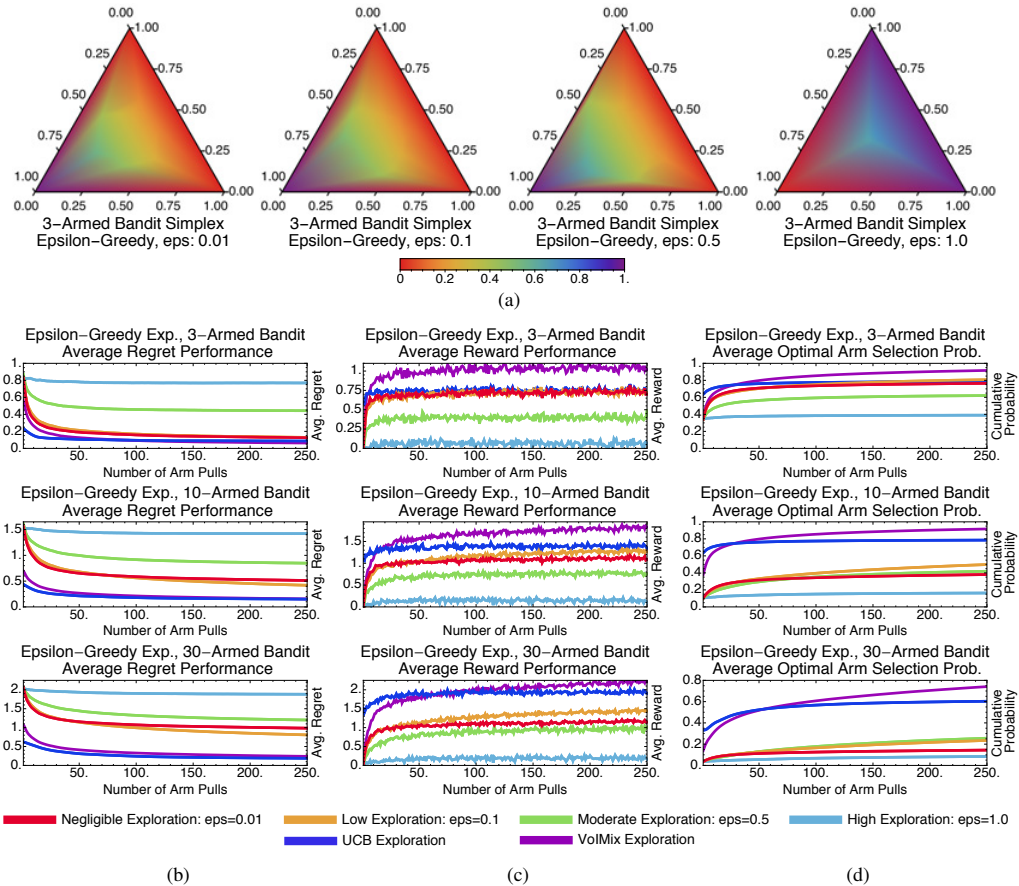

(a)

(b)                                          (c)                                          (d)

Figure 4.3: Results for the $\epsilon$-greedy-based GreedyMix for the fixed-parameter case. The plots in (a) highlight the regions in the action probability simplex that are visited during the simulations for the three-armed bandit problem. The correct action is the one associated with the bottom, left corner of the simplex. The simplex plots were produced for fixed parameter values and averaged across independent simulations. The plots in (b), (c), and (d) give the regret, reward, and optimal-arm selection cumulative probability for the three-, ten-, and thirty-armed bandit problem when the distribution variance is $\sigma^2 = 1$. The red, orange, green, and blue curves correspond to fixed exploration values of $\epsilon = 0.01$, $0.1$, $0.5$, and $1.0$, respectively.

### 4.3.1. Stochastic Methods: $\epsilon$-Greedy and Soft-Max

The $\epsilon$-greedy-based GreedyMix and soft-max-based SoftMix [20] are two of the more popular heuristic approaches for solving multi-armed bandits. GreedyMix is based upon the principle of taking random actions and random times and greedily choosing the best-paying arm in all other instances. SoftMix relies on a weighted-random approach to arm selection, similar to VoIMix and AutoVoIMix. That is, it weights each of the arms according to their expected pay-outs and stochastically chooses an arm using those weights.

Results for GreedyMix and SoftMix are given in figures 4.3 and 4.4, respectively. Despite only achieving a log-squared regret, SoftMix does well in the short term due to its weighted-random style of search. It accounts for the expected reward when deciding whether or not to explore new arms. Accurate means for well-paying arms are also created, which aids in choices that will initially reduce regret and improve rewards by a substantial amount; this can be seen in figures 4.3(b) and 4.3(c).

GreedyMix, in contrast, merely selects arbitrary arms, regardless of their quality. This haphazard selection can impede exploration early in the learning process, as sub-optimal arms may be frequently chosen. Poorer results than SoftMix are usually returned. This can be seen when comparing the plots from figures 4.3(b)–(d) to those from figures 4.4(b)–(d).

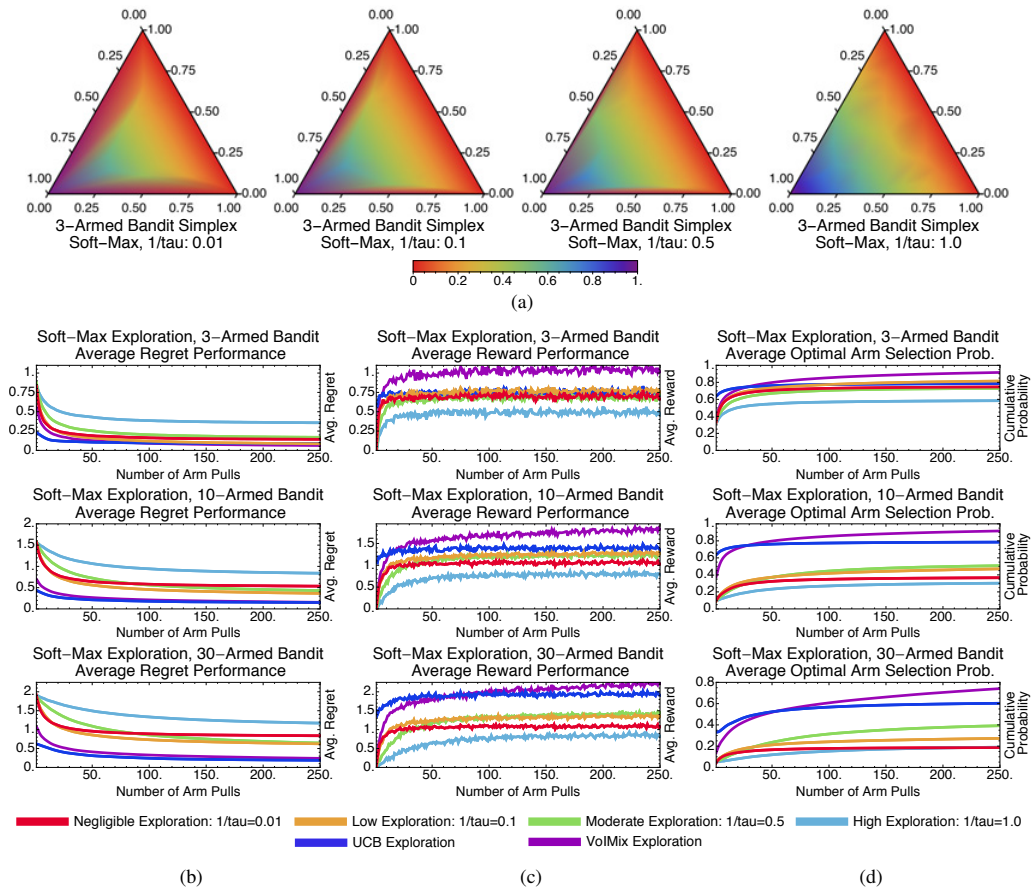

Figure 4.4: Results for the soft-max-based SoftMix for the fixed-parameter case. The plots in (a) highlight the regions in the action probability simplex that are visited during the simulations for the three-armed bandit problem. The correct action is the one associated with the bottom, left corner of the simplex. The simplex plots were produced for fixed parameter values and averaged across independent simulations. The plots in (b), (c), and (d) give the regret, reward, and optimal-arm selection cumulative probability for the three-, ten-, and thirty-armed bandit problem when the distribution variance is $\sigma^2 = 1$. The red, orange, green, and blue curves correspond to fixed inverse temperatures of $\tau^{-1} = 0.01$, $0.1$, $0.5$, and $1.0$, respectively.

Figures 4.3(a) and 4.4(a) illustrate that GreedyMix and SoftMix yield probability distributions that resemble those for the fixed-parameter VoIMix. Low amounts of exploration perform best for the three-armed bandit problem. The iterates tend to congregate in bottom-left corner of the simplex, which corresponds to the best-paying arm. The only exception is the right-most simplex plot in figure 4.3(a). The iterates continuously explore the entire simplex, since a new action is randomly chosen at each step.

Compared to VoIMix, both SoftMix and GreedyMix perform worse. GreedyMix will sometimes uniformly sample sub-optimal arms during the exploration phase. Its regret for the best-performing policy is thus higher, while the rewards are also reduced compared to the best results from VoIMix and AutoVoIMix. This is highlighted in figures 4.3(b)–(d) for VoIMix in the case where both the parameter and hyperparameter are adjusted across arm pulls. SoftMix also has issues. It has a habit of oversampling sub-optimal arms, which was caused by improperly estimated average rewards. Both the regret and total pay-out are worse, respectively, than VoIMix and AutoVoIMix. This is captured in figures 4.4(b)–(d).

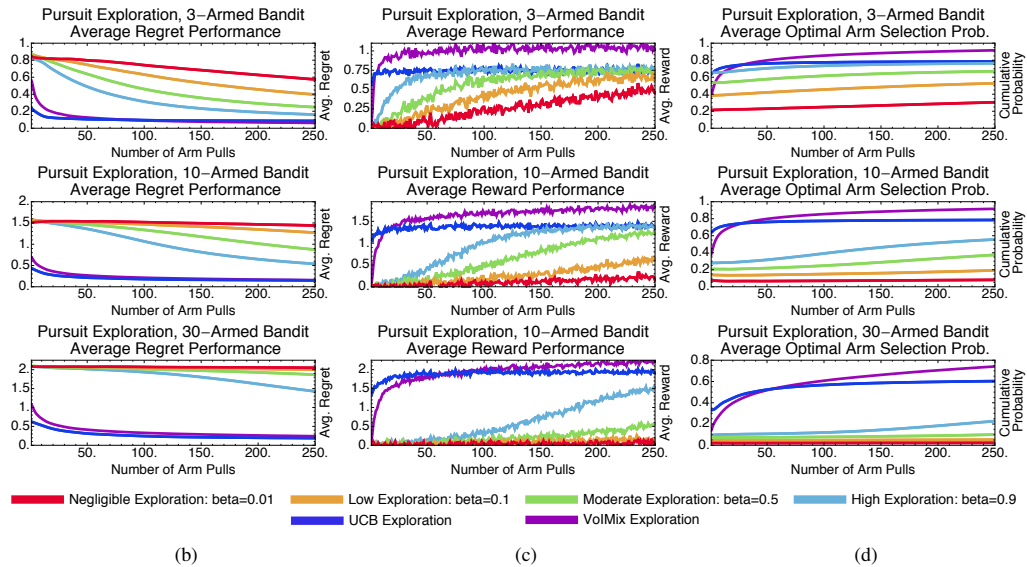

Figure 4.5: Results for the pursuit method for the fixed-parameter case. The plots in (a), (b), and (c) give the regret, reward, and optimal-arm selection cumulative probability for the three-, ten-, and thirty-armed bandit problem when the distribution variance is $\sigma^2 = 1$. The red, orange, green, and blue curves correspond to fixed values of the policy learning rate $\beta = 0.01, 0.1, 0.5$, and 0.9, respectively. Comparisons against the tuned version of VoIMix (purple curve), where $\tau_k^{-1}$ and $d_k$ change across each arm pull, along with the upper confidence bound method (dark blue curve) are provided.

### 4.3.2. Stochastic Methods: Pursuit and Reinforcement Comparison

The aforementioned stochastic exploration strategies are not the only ones available for addressing the multi-armed bandit problem. There are several others that can be applied, and we evaluate two of these, reinforcement comparison [1] and pursuit algorithms [38,39], both of which are nonparametric.

The methods that we have considered thus far are based on choosing arms according to their estimated expected value. Pursuit algorithms operate in a different fashion. Initially, they assign uniform probabilities to each arm. At each turn, more probability mass is given to the arm with the highest expected reward. Mass is removed from other arms, regardless of their estimated average rewards, at a rate determined by a learning factor. It has been demonstrated that such a scheme will converge, in a probably-approximately-correct sense, to the globally optimal policy. No formal regret bounds are currently available, though.

Results for the pursuit method are given in figure 4.5. Due to the way that exploration is performed, higher learning rates yielded better results than lower learning rates. Regardless of the learning rate, though, it is apparent that VoIMix produced sequences of arm pulls with better cumulative rewards and hence better regret. Likewise, it can be inferred, from figures 4.3 and 4.4, that both GreedyMix and SoftMix outperformed the pursuit method in the short term.

Reinforcement comparison methods are similar to pursuit methods, in that they maintain distributions over actions that are not directly derived from the empirical means. The way that they update these distributions sidesteps the biased-sampling issue. In particular, reinforcement comparison employs a two-step process. First, an estimate of the average pay-out for all slot machines is constructed. Second, more probability mass is given to those machines whose rewards exceed the overall average; machines that underperform have mass removed. The assignment of probability mass is based upon a Gibbs distribution. Previous analyses have revealed that this two-step process is guaranteed to converge to the optimal policy. However, no regret bounds have been made available as of yet.

Figure 4.6 presents the findings for reinforcement comparison. The plots show that this type of approach trailed modestly behind the tuned version of VoIMix. The time-averaged regret for reinforcement comparison is relatively high in the beginning and catches up with that from VoIMix. This lag is caused by the need to reliably estimate the expected reward for all machines. Over many episodes, the regret from VoIMix will be almost zero, as we showed in figure 4.2. The regret for reinforcement comparison stalls after a few hundred episodes; it does not appear to reach approximately zero regret on the same time scale as VoIMix.

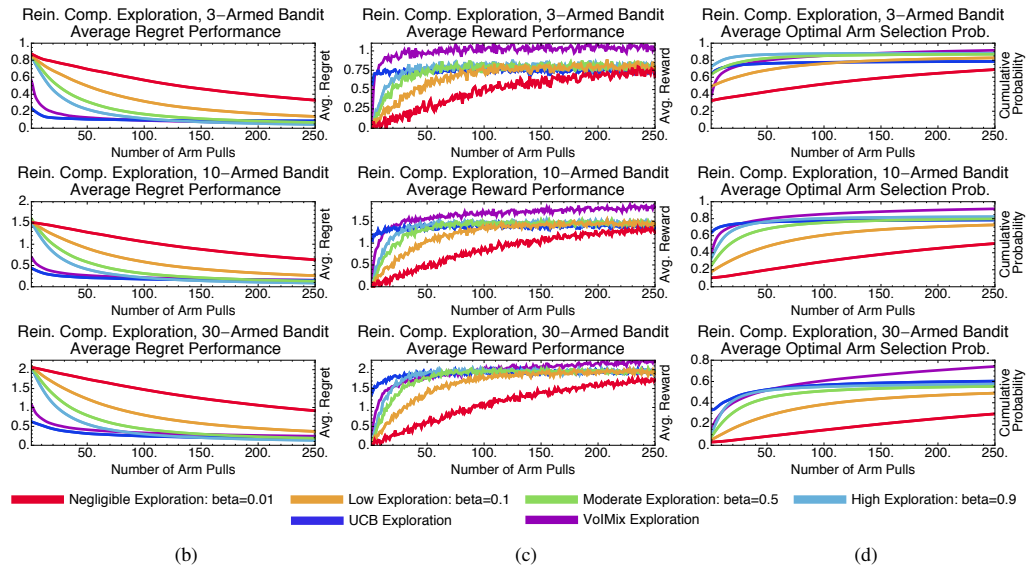

Figure 4.6: Results for the soft-max-based reinforcement comparison for the fixed-parameter case. The plots in (a), (b), and (c) give the regret, reward, and optimal-arm selection cumulative probability for the three-, ten-, and thirty-armed bandit problem when the distribution variance is $\sigma^2 = 1$. A value of $\alpha = 0.5$ was used for updating the mean reward at each iteration. The red, orange, green, and blue curves correspond to fixed values of the learning rate $\beta = 0.01, 0.1, 0.5$, and $0.9$, respectively. Comparisons against the tuned version of VoIMix (purple curve), where $\tau_k^{-1}$ and $d_k$ change across each arm pull, along with the upper confidence bound method (dark blue curve) are provided.

### 4.3.3. Deterministic Methods: Upper Confidence Bound

One of the most widely studied strategies for addressing multi-armed bandits are the deterministic upper-confidence-bound schemes [6,7]. They operate under the principle of optimism in the face of uncertainty. This entails choosing actions under the assumption that the unknown mean pay-offs of each arm is as large as plausibly possible, based on the previously observed samples.

There are many versions of the upper-confidence-bound idea. The simplest of these starts by pulling each arm once, from which it begins to construct an estimate of the arms' empirical means. It then greedily picks the arm with the highest pay-out, subject to an arm-frequency constraint [13]. This constraint ensures that arms that have not been played often, perhaps due to poorly estimated means, will eventually be selected. A permutation of this idea, called tuned upper confidence bound [40], takes into account both the means and the variance of the arm pay-outs, which purportedly aids in dealing with high-variance slot machines. Most, if not all, upper-confidence-bound approaches are known to achieve logarithmic regret without the need for parameter tuning.

The un-tuned upper-confidence-bound method results are presented in figures 4.3–4.6. The tuned variant produced virtually identical results for these problems, so those results are not shown in these plots. Both methods generated reasonably good time-averaged regret, per round, at the beginning of the simulations. In many cases, their initial performance exceeded that of the tuned VoIMix. The upper-confidence-bound methods also outperformed the remaining stochastic exploration approaches. They did converge much more slowly to the optimal arm than VoIMix, though. This led to lower pay-outs over a much larger set of plays.

### 4.3.4. Methodological Comparisons Discussions

These results indicate that the tuned version of VoIMix outperforms many conventional exploration schemes in the short term. Long-term it does too. We did not compare the results to the tuned version of AutoVoIMix, as its initial exploration period was longer than the number of pulls that we considered in these simulations; its results would be quite poor during this time. Over a greater number of arm pulls, the performance of AutoVoIMix approaches that of VoIMix. It will, consequently, yield better results than some of these alternate schemes.

The performance that we witnessed for many of these algorithms is predicated on choosing reasonable parameter or hyperparameter values. The average increase in total regret for improperly tuned algorithms was approximately thirty percent for VoIMix and forty percent for SoftMix and GreedyMix. Reinforcement comparison likewise suffered an increase in regret of about forty percent, as did the pursuit method, which saw a seventy percent raise. Moreover, seemingly optimal parameter values for one multi-armed bandit scenario could suddenly become one of the worst in another. This finding indicates a need for online parameter adjustment, akin to what is done by the information-theoretic scheme that we introduced for adapting the value-of-information hyperparameter.

Below, we discuss aspects of the obtained simulation results.

$\epsilon$-**Greedy and Soft-Max Discussions.** Our simulations indicated that the performance of SoftMix exceeded that of GreedyMix. In the long-term, GreedyMix should improve beyond SoftMix, especially if the currently known regret bounds are tight. This is because an unlucky sampling of arms in SoftMix can bias the expected returns and favor sub-optimal arms. The optimal arm may have a low probability of being sampled, especially after many pulls. In contrast, all arms have the equal chance of being selected during the exploration phase of GreedyMix. Any poorly estimated reward means, especially that of the optimal arm, can hence be made more precise. The optimal arm will therefore be chosen more frequently once this occurs.

Despite both methods achieving a logarithmic regret, VoIMix outperformed the fixed-parameter GreedyMix over the first few hundred arm pulls. This was also observed in the case where the parameters for GreedyMix were annealed across arm pulls. This discrepancy is likely due to the constant regret complexity factors, which are typically excluded. Those factors for VoIMix have a lower magnitude than those for GreedyMix.

**Pursuit and Reinforcement Comparison Discussions.** Both VoIMix and AutoVoIMix perform well since they avoid a biased sampling of sub-optimal arms. Pursuit methods, in contrast, are highly susceptible to such issues. These types of algorithms only consider the expected reward associated with the current-best arm when assigning probability mass. The rewards of all other arms are ignored, unlike in soft-max-based selection, which provides little opportunity to recover from a poor arm selection after a few iterations. Likewise, the uniform-exploration components in VoIMix and AutoVoIMix further aid in avoiding a biased sampling, especially early in the learning process where it dominates. We found that low-reward policies were formulated rather frequently, in practice, by pursuit methods. The regret tended to plateau to a sub-optimal value in just a few arm pulls. The optimal arm was rarely identified, even after adjusting the learning-rate parameter.

Reinforcement comparison behaved comparably to both GreedyMix and SoftMix for the problems that we considered. In high-variance cases, it sometimes excelled beyond these two approaches. Such behavior is largely due to its mean-discrimination capabilities. That is, reinforcement comparison would typically waste only a few pulls on low-mean arms that had unusually high rewards for a few rounds. Once a better estimate of the expected reward began to form these sub-optimal arms, it would quickly switch to others. Substantial parameter tuning was needed for this shift to happen without greatly impacting regret, though. This contrasts with approaches like VoIMix, where good performance was obtained without significant experimentation.

In short, as noted by Cesa-Bianchi, et al. [20,21], and echoed by several other authors in the literature, finding good annealing schedules for stochastic methods can be difficult. Some cooling schedules may consistently draw suboptimal arms during the learning process, even after having estimated all of the means correctly. In some cases, they may focus too early to a suboptimal arm and never recover afterwards. The value of information appears to avoid such issues, due to its mixture-model formulation. That is, there is always a chance that a non-optimal arm can be selected uniformly at random, even if its current reward estimate is incorrect.

**Upper-Confidence-Bound Discussions.** Our simulations revealed that while upper-confidence-bound methods achieve superior performance during the first hundred arm pulls, it lags behind VoIMix over a longer number of plays. This behavior likely stems from the greedy arm-selection process of the upper-confidence-bound methods. Initially, only the highest-performing slot machine will be chosen by these methods, as the arm-frequency constraint will do little to sway the selection. This arm may be sub-optimal, yet have an unusually high reward due to the variance. The optimal arm, in contrast, may have had an uncharacteristically low reward the round that it was chosen, causing it to be ignored until later in the learning phase when the arm-frequency constraint will ensure that it is picked. Stochastic approaches have more opportunities to choose optimal arms that, initially, appear to be sub-optimal.

Both the tuned and un-tuned upper-confidence-bound methods yielded near-identical regrets and rewards in our simulations. Increasing the arm variance did little to change the results. The pay-outs accrued by each of the algorithms that we consider thus appear to only be affected by two characteristics. These are the number of slot-machine arms and the mean separation of the corresponding machine distributions.

This finding has ramifications for the types of regret bounds that investigators can expect to obtain. It also informs about the methodological traits that should be emphasized when developing new bandit algorithms. Such a suggestion runs counter to recent theoretical endeavors that have focused on obtaining improved regret bounds by considering the second-order reward moments of specific distributions. Nevertheless, our results indicate that accounting for high-order moments of the reward distribution will not prove fruitful, as the type of reward distribution had only a marginal effect on the algorithm behaviors and performance. This is a trait exhibited not only by the value of information, but also for the other stochastic and deterministic approaches that we considered here.

## 5. Conclusions

In this paper, we considered the exploration of discrete, stochastic multi-armed bandits. This simplified reinforcement learning problem is important, as it is often difficult to assess certain performance aspects of learning methods for multi-state, multi-action Markov decision processes.

We proposed an information-theoretic means of choosing optimal actions in a single-state, multi-action environment. Our approach is based on a value of information criterion, which defines a type of free-energy extremization problem. We have previously empirically analyzed this criterion for multi-action, multi-state reinforcement learning. Here, however, we focused on a more theoretical treatment, which is a contribution of our paper. The value of information has not been analyzed for single-state, multi-action environments before, to our knowledge, so this contribution offers insights into the search capabilities of this criterion.

Optimization of the value-of-information criterion gives rise to a soft-max-like policy search. Exploration is therefore driven by weighted randomness. The weights are determined by a parameterized Gibbs distribution, which operates on the rewards associated with pulling various arms. The parameter, a type of inverse temperature, dictates the amount of exploration that is conducted in the policy space.

A purely soft-max-based action selection suffers from a well-known flaw. That is, an infinite number adjustments of the inverse temperature parameter are needed to convert from a purely random exploration of the policy space to pure exploration. To resolve this issue, we created a mixture model from the Gibbs distribution that can more effectively transition from one extreme to the other. This model defines the action-selection probabilities to be a mixture of the exponential-based Gibbs-distribution component and a uniform-distribution component. The influence of either of these components is determined by a single mixing parameter.

Both mixture-model parameters have a profound impact on the accrued rewards. We considered two possible means of automatically tuning the parameters during the search process. The first approach, VoIMix, is optimal, as it achieves logarithmic regret for the multi-armed bandit problem. It does, however, rely on a user-specified term that is difficult to set since it relies on details about the problem that are not normally available a priori. This limits its practical use. The second approach, AutoVoIMix, is nearly optimal, as it can be arbitrarily close to logarithmic regret. The user-selectable term is much easier to set for this case, though, as it does not depend on any a priori unknown quantities.

We assessed the search capability of the value of information through a battery of simulations. This is another contribution of our paper. When manually choosing the amount of exploration, VoIMix either performs similarly or outperforms conventional heuristics, such as $\epsilon$-greedy and soft-max selection, in terms of regret and overall reward. When annealing the exploration amount, both VoIMix and AutoVoIMix outperformed these criteria. This was expected. $\epsilon$-greedy does not account for the reward associated with action choices. It is therefore possible for poor rounds of initial exploration to occur, which limits the short-term obtainable rewards. This occurs whether the amount of exploration is either manually fixed or automatically chosen so as to achieve logarithmic regret. Soft-max selection, in contrast, is not known to exhibit logarithmic regret. Existing implementations of soft-max selection also have the flaw that the consistent sampling some sub-optimal arm may prevent the optimal arm from being sampled enough.

In the limit, the value of information is guaranteed to reach zero regret with unit probability. Many other bandit algorithms share this trait. To help our approach reach a near-zero regret more quickly, we relied on a cross-entropy heuristic for monitoring the evolution of the search and updating the model hyperparameters. This heuristic can be applied to more general reinforcement learning problems too. Our application of this heuristic to the multi-armed bandit problem is another contribution of our paper.

We also analyzed the performance of the tuned and untuned versions of VoIMix against AutoVoIMix. The former leads to steep drops in regret due to an effective action exploration, provided that the parameter values are in the valid range. The regret for the tuned VoIMix approach is often better than the best fixed-exploration-amount cases after only a few hundred arm pulls. The optimal arms are routinely identified with a probability near one. AutoVoIMix, in contrast, has the tendency to unnecessarily search the action space. A great many arm pulls are needed for the regret to become small, even when its hyperparameters are tuned according to our cross-entropy heuristic. While both VoIMix and AutoVoIMix can achieve the same regret and instantaneous pay-out in the limit, the former would be poorly suited for real-world, time-sensitive bandit problems where few arm pulls are made before learning stops.

Our theoretical analyses indicate that an adaptation of the exploration rate is needed for the value of information to achieve logarithmic regret. This was corroborated by our simulation results: fixed amounts of exploration led to poorer results. We have witnessed similar behaviors for the multi-state, multi-action case. In our future work, we will ascertain if any of the theory from this paper can be applied to the multi-state case so that the optimal average cost improvement can be obtained per episode.

Certain logarithmic-regret approaches, such as extensions of the upper-confidence-bound algorithm, have explicit constant terms. It can be shown that the constant terms for some of these approaches are superior to that of VoIMix and AutoVoIMix, indicating that the former will achieve better asymptotic performance than our methods. Another area of future research will hence be a refinement of the constant term in the logarithmic regret bound that we obtained for VoIMix. Toward this end, we have already started combining the value of information with the upper confidence bound method. The outcome of this combination is that the number of simulated transitions made to arrive at an optimal policy will be reduced, which supplies a better constant term in the regret bound. While our simulation results here have indicated that VoIMix is competitive against state-of-the-art techniques, the performance will likely improve further over finite numbers of plays due to this algorithmic change.

## Appendix A

In this section, we provide a proof of the logarithmic regret for algorithm 1. We define the (expected) regret after the first $k$ pulls, according to some policy that chooses a sequence of arms $i_1, \ldots, i_k$, to be

$$\max_{1 \leq j \leq |\mathcal{A}|} \mathbb{E}\left[\sum_{s=1}^{k} X_s^j - X_s^{i_s}\right],$$

where $X_s^{i_s}$ represents the reward obtained by pulling arm $a_s^i$ at round $s$. Here, the expectation operator is with respect to the stochastic generation of rewards. For randomized rules, it is also with respect to the internal randomization of the rule.

In the standard formulation of the stochastic, multi-armed bandit problem, it is assumed that the rewards obtained from pulling arms are independent random variables $X_k^i$ with stationary means $\mu^i$, for all $i, k$. All of the following results hold under this assumption. Our results also are valid under a weaker assumption that $\mathbb{E}[X_k^i | \mathcal{J}_{k-1}] = \mu^i$, for all $i, k$, where $\mathcal{J}_{k-1}$ represents the $\sigma$-field generated by the random variables. This weaker assumption states that the distribution of each new random reward can depend, in an adversarial way, on the previous pulls and observed rewards, provided the reward mean is fixed. That is, the reward distributions can change in an adversarial way according to some external process [4]. The bounds hence apply to the adversarial bandit algorithm without modification.

Our proof of logarithmic regret hinges on showing that we can obtain, for $\mathbb{E}[\pi_k^i]$, an upper bound of $O(1/k)$ for any sub-optimal slot-machine arm. Here, $\pi_k^i$ is the probability that $I_{k+1}^i = 1$ for some sub-optimal arm. From this result, logarithmic regret easily follows, since the sum of the difference in optimal and sub-optimal reward means for each arm is equal to $O(\log(k))$.

It is important to note that the expected value of $\pi_k^i$ can be obtained after any number $k$ of arm pulls. This is a distinction compared to other regret results in the literature, which are valid only for $k$ going to infinity.

We first provide a meaningful intermediate bound that will be useful throughout our subsequent analyses.

**Proposition A.1.** For algorithm 1, we have, for sub-optimal arms $a^i \in \mathcal{A}$ at iteration $k$, that

$$\mathbb{E}[\pi_k^i] \le \exp(\tau_k^{-1}\delta_i - k\tau_k^{-1}\delta_i)\mathbb{E}\left[\prod_{s=1}^{k-1}\exp\left(\phi_c(\tau_s^{-1})\left(2\frac{|\mathcal{A}|}{\gamma_s} - \delta_i^2\right)\right)\right] + \frac{\gamma_k}{|\mathcal{A}|}.$$

Here, $\delta_i$ is the difference in expected pay-outs, $\delta_i = \mu^* - \mu^i$, for the optimal $a^*$ versus non-optimal $a^i$ arms. The function $\phi_c(\tau_s^{-1})$ is given by $\phi_c(\tau_s^{-1}) = (e^{c\tau_s^{-1}} - c\tau_s^{-1} - 1)/\tau_s^{-2}$.

**Proof:** We introduce a random variable $V_p^j = X_p^j/\pi_p^j$ if $I_p = j$ and zero otherwise. It can be seen that $V_p^j \le 1/\pi_p^j$ and that $1/\pi_p^j \le |\mathcal{A}|/\gamma_p$. Both of these facts are implicitly used.

In what follows, we systematically bound the expected value of the policy. Toward this end, we note that the following relationships hold for $\mathbb{E}[\pi_k^i]$,

$$(1 - \gamma_k)\mathbb{E}\left[\frac{\exp(\sum_{s=1}^{k-1}\tau_s^{-1}V_s^i)}{\sum_{j=1}^{|\mathcal{A}|}\exp(\sum_{s=1}^{k}\tau_s^{-1}V_s^j)}\right] + \frac{\gamma_k}{|\mathcal{A}|} \le (1 - \gamma_k)\mathbb{E}\left[\frac{\exp(\sum_{s=1}^{k-1}\tau_s^{-1}V_s^i)}{\exp(\sum_{s=1}^{k}\tau_s^{-1}V_s^*)}\right] + \frac{\gamma_k}{|\mathcal{A}|}$$

$$= (1 - \gamma_k)\mathbb{E}\left[\exp\left(\sum_{s=1}^{k-1}\tau_s^{-1}V_s^i - \tau_s^{-1}V_s^*\right)\right] + \frac{\gamma_k}{|\mathcal{A}|}$$

where the inequality is due to the definition of the random variable $V_s^j$. The equality follows from the fact that the fractional exponential terms inside the expectation can be combined into a single expression. We now refine this bound so that we can eventually show that it is the reciprocal of the number of arm pulls. We begin by factoring out the summation of $\tau_s^{-1}V_s^i - \tau_s^{-1}V_s^*$ terms inside the exponential function. Continuing from above,

$$= (1 - \gamma_k)\mathbb{E}\left[\prod_{s=1}^{k-1}\mathbb{E}\left[\exp(\tau_s^{-1}V_s^i - \tau_s^{-1}V_s^*)\Big|\mathcal{J}_{s-1}\right]\right] + \frac{\gamma_k}{|\mathcal{A}|}$$

$$\le \exp(\tau_k^{-1}\delta_i - k\tau_k^{-1}\delta_i)\mathbb{E}\left[\prod_{s=1}^{k-1}\mathbb{E}\left[\exp(\tau_s^{-1}Z_s^i)\Big|\mathcal{J}_{s-1}\right]\right] + \frac{\gamma_k}{|\mathcal{A}|}.$$

In the last step, we multiplied and divided by an exponential quantity $\exp(k\tau_k^{-1}(\mu^j - \mu^*) - \tau_k^{-1}(\mu^j - \mu^*))$. We also dropped the multiplicative factor $1 - \gamma_k$. In this expression, $Z_s^i = \delta_i + V_s^i - V_s^*$ is another random variable.

To further refine the bound for $\mathbb{E}[\pi_p^i]$, we need to deal with the factor $\mathbb{E}[\exp(\tau_s^{-1}Z_s^i)|\mathcal{J}_{s-1}]$; here, $\mathcal{J}_{s-1}$ represents a $\sigma$-algebra. We will rely on a Taylor expansion of the exponential function to do this. This results in the expression $\mathbb{E}[\exp(\tau_s^{-1} Z_s^i)|\mathcal{J}_{s-1}] \leq \mathbb{E}[1 + \tau_s^{-1} Z_s^i + Z_s^i Z_s^i \phi_c(\tau_s^{-1})|\mathcal{J}_{s-1}]$, each term of which can be dealt with independently.

We first note that the conditional expectation, for each $s$, is given by the following expression $\mathbb{E}[Z_s^i|\mathcal{J}_{s-1}] = \mathbb{E}[V_s^i|\mathcal{J}_{s-1}] - \mathbb{E}[V_s^*|\mathcal{J}_{s-1}] + \delta_i$, which is zero. Next, we bound the conditional variance, for each $s$, $\mathbb{E}[Z_s^i Z_s^i|\mathcal{J}_{s-1}]$. It can be seen that $\mathbb{E}[Z_s^i Z_s^i|\mathcal{J}_{s-1}] = \mathbb{E}[(V_s^i - V_s^*)^2|\mathcal{J}_{s-1}] - \delta_i^2$. Expanding the squared term, we find that the conditional variance is equal to the sum of four factors. The factor $\mathbb{E}[V_s^* V_s^*|\mathcal{J}_{s-1}] - 2\mathbb{E}[V_s^* V_s^i|\mathcal{J}_{s-1}]$ can be dropped. After expanding the remaining expectation terms, we find that

$$\mathbb{E}[\pi_k^i] \leq \exp(\tau_1^{-1}\delta_i - k\tau_k^{-1}\delta_i)\mathbb{E}\left[\prod_{s=1}^{k-1} \mathbb{E}\left[1 + \tau_s^{-1} Z_s^i + Z_s^i Z_s^i \phi_c(\tau_s^{-1})\Big|\mathcal{J}_{s-1}\right]\right] + \frac{\gamma_k}{|\mathcal{A}|}$$

$$\leq \exp(\tau_1^{-1}\delta_i - k\tau_k^{-1}\delta_i)\mathbb{E}\left[\prod_{s=1}^{k-1}\left(\mathbb{E}\left[\frac{X_s^i X_s^i}{\pi_s^i \pi_s^i}\pi_s^i\Big|\mathcal{J}_{s-1}\right] + \mathbb{E}\left[\frac{X_s^* X_s^*}{\pi_s^* \pi_s^*}\pi_s^*\Big|\mathcal{J}_{s-1}\right] - \delta_i^2\right)\right] + \frac{\gamma_k}{|\mathcal{A}|}$$

$$\leq \exp(\tau_1^{-1}\delta_i - k\tau_k^{-1}\delta_i)\mathbb{E}\left[\prod_{s=1}^{k-1}\exp\left(\phi_c(\tau_s^{-1})\left(2\frac{|\mathcal{A}|}{\gamma_s} - \delta_i^2\right)\right)\right] + \frac{\gamma_k}{|\mathcal{A}|}.$$

which is the desired bound. ∎

We now show that algorithm 1 achieves logarithmic regret for the first choice of $\tau_k^{-1}$. A proof for the second choice of $\tau_k^{-1}$ in algorithm 1 is similar and therefore omitted.

**Proposition A.2.** The value-of-information update in algorithm 1 yields $\mathbb{E}[\pi_k^i] = O(1/k)$ for sub-optimal arms $a^i \in \mathcal{A}$ at iteration $k$. This holds under the assumption that $\gamma_k = \min(1, 5|\mathcal{A}|/kd^2)$ and $\tau_k^{-1} = \log(1 + dc_k/(c_k-1))/c_k$ with $c_k = 2|\mathcal{A}|/\gamma_k + 1$. It also holds when using $\tau_k^{-1} = \log(1 + dc_k/\sigma_k^2)/c_k$, with $\sigma_k^2 = 2|\mathcal{A}|/\gamma_k - d^2$, to iteratively update the inverse-temperature parameter.

**Proof:** We have that

$$\mathbb{E}[\pi_k^i] \leq (1 - \gamma_k)\exp\left(-\sum_{s=1}^{k-1}\tau_s^{-1} d - \phi_c(\tau_s^{-1})\sigma_s^2\right) + \frac{\gamma_k}{|\mathcal{A}|}$$

where $\sigma_s^2 = 2|\mathcal{A}|/\gamma_s - d^2$. We now write $\tau_s^{-1} = \log(1 + k_s c_s/\sigma_s^2)/c_s$, where $k_s = d$. We apply an elementary logarithmic identity to bound $\tau_s^{-1}$: $\log(1 + k_s c_s/\sigma_s^2)c_s \geq 2k_s/(2\sigma_s^2 + k_s c_s)$. Therefore,

$$\mathbb{E}[\pi_k^i] \leq (1 - \gamma_k)\exp\left(-\sum_{s=1}^{k-1}\tau_s^{-1} k_s - \left(\frac{\exp(c_s\tau_s^{-1}) - 1}{c_s^2} - \frac{2k_s}{c_s(2\sigma_s^2 + k_s c_s)}\right)\sigma_s^2\right) + \frac{\gamma_k}{|\mathcal{A}|}$$

$$\leq \exp\left(-\sum_{s=1}^{k-1}\frac{k_s^2}{2\sigma_s^2 + k_s c_s}\right) + \frac{\gamma_k}{|\mathcal{A}|}$$

where $\exp(c_s\tau_s^{-1} - 1) = k_s c_s/\sigma_s^2$ due to the fact that $\tau_s^{-1} = \log(1 + k_s c_s/\sigma_s^2)/c_s$. Substituting the variables $c_s$, $k_s$, and $\sigma_s^2$, we find that $k_s^2/(2\sigma_s^2 + k_s c_s) \geq d^2\gamma_s/5|\mathcal{A}|$. The factor of $5|\mathcal{A}|$ arises as a lower bound for the denominator of $k_s^2/(2\sigma_s^2 + k_s c_s)$ since $2\sigma_s^2 + k_s c_s = 4|\mathcal{A}| + d|\mathcal{A}| + \gamma_s(d - 2d^2)$. We therefore obtain that $\mathbb{E}[\pi_k^i] \leq q\exp(-\log(k-1)) + d^2\gamma_s/5|\mathcal{A}|$, where the constant $q$ represents the regret accumulated for indices $s$ up to a finite point. It is immediate that $\mathbb{E}[\pi_k^i] \leq q/(k-1) + d^2\gamma_s/5|\mathcal{A}|$ after canceling out the logarithmic and exponential functions. ∎

For the second choice of revising the inverse temperature $\tau_k^{-1}$ in algorithm 1, we can provide non-averaged bounds on the probability $\pi_k^i$ of choosing a sub-optimal arm $a^i$. This result shows just how much $\pi_k^i$ can change across iterations.

**Proposition A.3.** For sub-optimal arms $a^i \in \mathcal{A}$ at iteration $k$, of the value-of-information update in algorithm 1 yields $\gamma_k / |\mathcal{A}| \leq \pi_k^i \leq \gamma_k / |\mathcal{A}| + (1 - \gamma_k)/k$ with probability $1 - k(\omega')^k$ for $0 < \omega < 1$. This holds under the assumption that $\gamma_k = \min(1, 5|\mathcal{A}|/kd^2)$ and $\tau_k^{-1} = \log(1 + dc_k/(c_k - 1))/c_k$, with $c_k = 2|\mathcal{A}|/\gamma_k + 1$, are used to update the mixing coefficient and inverse-temperature parameters.

   **Proof:** The bound $\gamma_k / |\mathcal{A}| \leq \pi_k^i$ on the left-hand side is evident. It is simply the exploration probability for arm $a^i \in \mathcal{A}$ when drawn from the uniform distribution term.

The bound $\pi_k^i \leq \gamma_k / |\mathcal{A}| + (1 - \gamma_k)/k$ on the right-hand side follows from our above results and various probabilistic inequalities. We first note that, for $\pi_k^i$, the following holds

$$(1 - \gamma_k) \left( \frac{\exp(\sum_{s=1}^{k-1} \tau_s^{-1} V_s^i)}{\sum_{j=1}^{|\mathcal{A}|} \exp(\sum_{s=1}^{k} \tau_s^{-1} V_s^j)} \right) + \frac{\gamma_k}{|\mathcal{A}|} \leq (1 - \gamma_k) \left( \exp\left( \sum_{s=1}^{k-1} \tau_s^{-1} V_s^i - \tau_s^{-1} V_s^* \right) \right) + \frac{\gamma_k}{|\mathcal{A}|}$$

where we have defined the random variable $V_k^j = X_k^j / \pi_k^j$ if $I_k^j = 1$ and zero otherwise. Using the Markov inequality, we obtain that

$$(1 - \gamma_k) p \left( \exp\left( \sum_{s=1}^{k-1} \tau_s^{-1} V_s^i - \tau_s^{-1} V_s^* > \frac{1}{k} \right) \right) + \frac{\gamma_k}{|\mathcal{A}|} \leq k \mathbb{E}\left[ \sum_{s=1}^{k-1} \mathbb{E}\left[ \exp(\tau_s^{-1} Z_s^i) \Big| \mathcal{J}_{s-1} \right] \right] + \frac{\gamma_k}{|\mathcal{A}|}$$

where $Z_k^i = V_k^i - V_k^*$ is a random variable used to describe the difference in probabilistically-weighted rewards.

We introduce another random variable $Q_{k-1}^i = \prod_{s=1}^{k-1} \exp(\tau_s^{-1} V_s^i - \tau_s^{-1} V_s^*)$ to model the product of exponential terms. Observe that $Q_{k-1}^i = (V_{k-1}^i - V_{k-1}^*) Q_{k-2}^i = Z_{k-1}^i Q_{k-2}^i$. Also note that $\mathbb{E}[Q_{k-1}^i] = \mathbb{E}[Q_{k-1}^i | \mathcal{J}_{k-2}]$, where $\mathcal{J}_{k-1}$ is $\sigma(V_p^i, 1 \leq p \leq k-2)$ with $k \geq 3$.

From what we derived above, we have $\mathbb{E}[Z_{k-1}^i | \mathcal{J}_{k-2}] = -\delta_i$. We also have $\mathbb{E}[Z_{k-1}^i Z_{k-1}^i | \mathcal{J}_{k-2}] \leq 2|\mathcal{A}|/\gamma_{k-1}$. It can be seen that $Z_{k-1}^i \leq |\mathcal{A}|/\gamma_{k-1}$ and hence that $Z_{k-1}^i / c_{k-1} \leq 1/2$. Lastly, $|\tau_{k-1} Z_{k-1}^i| \leq 1$ since $\log(1 + dc_{k-1}/(c_{k-1}-1)) = \log(1 + d + d\gamma_{k-1}/2|\mathcal{A}|) < 1$ for small values of $d$.

We thus get that $\mathbb{E}[Q_{k-1}^i] \leq (1 - \tau_{k-1}^{-1} \delta_i + \tau_{k-1}^{-2} 2|\mathcal{A}|/\gamma_{k-1}) Q_{k-1}^i$ and

$$p\left( \exp\left( \sum_{s=1}^{k-1} \tau_s^{-1} V_s^i - \tau_s^{-1} V_s^* > \frac{1}{k} \right) \right) \leq k \mathbb{E}\left[ \sum_{s=1}^{k-1} \mathbb{E}\left[ 1 + \tau_s^{-1} Z_s^i + \tau_s^{-2} Z_s^i Z_s^i \Big| \mathcal{J}_{s-1} \right] \right]$$

$$\leq k \left( 1 - \tau_{k-1}^{-1} \delta_i + \tau_{k-1}^{-2} \frac{2|\mathcal{A}|}{\gamma_{k-1}} \right) \prod_{s=1}^{k-1} \exp(\tau_s^{-1} V_s^i - \tau_s^{-1} V_s^*).$$

The multiplicative factor $\tau_{k-1}^{-1} d_i - \tau_{k-1}^{-2} 2|\mathcal{A}|/\gamma_{k-1}$ can be bounded below as follows $\tau_{k-1}^{-1} d_i - \tau_{k-1}^{-2} 2|\mathcal{A}|/\gamma_{k-1} \geq d\tau_{k-1}^{-1} - d\tau_{k-1}^{-1}/(1 + dc_{k-1}/\sigma k - 1^2)^{1/2} \geq \omega'$, where $\omega' > 0$. This follows from the fact that $2\tau_{k-1}^{-1} |\mathcal{A}|/\gamma_{k-1}$ is equal to $\sigma_{k-1}^2 \log(1 + dc_{k-1}/\sigma_{k-1}^2)/c_{k-1}$, where $\sigma_{k-1}^2 / c_{k-1} < 1$, along with an application of a logarithmic inequality. Setting $\omega = 1 - \omega'$, we arrive at $\mathbb{E}[Q_k^i] \leq (\omega')^k Q_{k-1}^i$, where $(\omega)^k$ denotes taking the $k$th power of $\omega$. Therefore, we obtain that the probability is $p(\exp(\sum_{s=1}^{k-1} \tau_s^{-1} Z_s^i) > 1/k) < k(\omega')^k$.                ∎

The preceding result can be employed to illustrate the competitiveness of the value of information against other schemes. For example, Auer et al. [13] supplied a probability bound of choosing a non-optimal slot machine arm, after $k \geq c|\mathcal{A}|/d$ pulls, when using only a uniform-exploration component. This upper bound can be approximated by $O(c/d^2k) + O(1/k)$ for $k$ large. We, however, have obtained a bound of $O(d^2k/k) + O(1/d^2k)$, which is strictly less than the one provided by Auer et al. Value-of-information-based search hence has a higher chance of finding an optimal arm much sooner than pure $\epsilon$-greedy schemes, which explains the experimental results that we obtained.

This result additionally indicates that the exponential component in the arm-selection update is important for resolving the exploration-exploitation dilemma. It can be shown that much of the exploration is driven by the uniform-random component of the mixture model. The role of the exponential, Gibbs-based component is to force the switch to exploitation much more quickly than in $\epsilon$-greedy-based approaches like [13]. This permits the highest-paying arm to be chosen more consistently earlier during training; this arm has a high chance of being the optimally paying one.

As we noted above, a principled guess for the parameter $d$ is needed, for VoIMix in algorithm 1, to adjust the temperature parameter. This guess implicitly requires knowledge of the unknown slot-machine expected values $\mu^j$, for $a^j \in \mathcal{A} \backslash a^*$, let alone the optimal expected value $\mu^*$ for $a^* \in \mathcal{A}$. Neither of these quantities can be reliably obtained a priori for arbitrary problems. For this reason, we introduced an alternative in algorithm 2, which provides a means of adjusting the temperature parameter with an easier-to-set term $\theta$.

We now show that AutoVoIMix, given in algorithm 2, can achieve logarithmic regret in the limit.

**Proposition A.4.** For the value-of-information update in algorithm 2, we have that $\mathbb{E}[\pi_k^i] = O(\log^{2\theta}(k)/k)$ for sub-optimal arms $a^i \in \mathcal{A}$ at iteration $k$. This holds under the assumption that $\gamma_k = \min(1, 5|\mathcal{A}|\log^{2\theta}(k)/k)$ and $\tau_k^{-1} = \log(1 + c_k \log^{-\theta}(k)/(2c_k - 2))/c_k$ with $c_k = |\mathcal{A}|/\gamma_k + 1$.

**Proof:** From propositions A.1 and A.2, it can be shown that

$$\mathbb{E}[\pi_k^i] \leq (1 - \gamma_k)\exp\left(-\sum_{s=1}^{k-1} \tau_s^{-1}k_s - \phi_c(\tau_s^{-1})\sigma_s^2\right) + \frac{\gamma_k}{|\mathcal{A}|}$$

$$\leq \exp\left(-\sum_{s=1}^{k-1} \frac{k_s^2}{2\sigma_s^2 + k_s c_s}\right) + \frac{\gamma_k}{|\mathcal{A}|}$$

for $k_s = \log^{-\theta}(s)$ and $\sigma_s^2 = 2|\mathcal{A}|/\gamma_s$. The summation terms $k_s^2/(2\sigma_s^2 + k_s c_s)$ can be further bounded by $\gamma_s \log^{-2\theta}(s)/5|\mathcal{A}|$. After substituting the expression for $\gamma_k$ in $\gamma_s \log^{-2\theta}(s)/5|\mathcal{A}|$, we find that $\mathbb{E}[\pi_k^i] = O(1/k)$. Plugging in the form of $\gamma_k$ in the exploration term $\gamma_k/|\mathcal{A}|$ yields $\mathbb{E}[\pi_k^i] = O(\log^{2\theta}(k)/k)$.　■

The expected regret when using this algorithm is $O(\log(k)^{1+2\theta})$. The regret can be made infinitesimally close to logarithmic by choosing a small, non-zero value of $\theta$. Note, however, values of $\theta$ have an inverse relationship with the number of arm pulls. For $\theta$ to be as small as possible, a great many arm pulls must be made to reach the expected regret bound.

The regret bound for AutoVoIMix is important for a few reasons. Foremost, it is valid for both stochastic and adversarial bandit problems. In the latter case, the rewards earned from the slot-machine can be a combination of any previously obtained rewards. This represents a more difficult problem than the stochastic bandit case, where the rewards are independent of each other.

Secondly, as we noted above, this regret bound is obtained independently of a hyperparameter that depends on statistics of the rewards. That is, the bound is distribution-free. There are other distribution-free methods in the literature, like Exp3 or Exp4 [24]. However, such methods typically only obtain linear-logarithmic regret: the regret of Exp3 is $O((|\mathcal{A}|k\log(|\mathcal{A}|k/\kappa))^{1/2})$, while the regret for Exp4 is $O((|\mathcal{A}|k\log(n))^{1/2})$, with $n$ being the number of expert strategies that are supplied. Both bounds are rather poor, which is due to the fact that no information about the difficulty of the search problem is being conveyed through the hyperparameter that adjusts the exploration amount. Both Exp3 and Exp4 hence require a large amount of exploration regardless of if the optimal-to-suboptimal slot-machine average payout difference is large or small. AutoVoIMix, in contrast, requires less exploration than either Exp3 or Exp4. This marked decrease in exploration is primarily due to the gradual modification of the exploration hyperparameters across each arm pull of a given simulation. For Exp3 and Exp4, this hyperparameter remains fixed throughout a given simulation and is only modified across simulations for certain practical implementations.

## References

1.　Sutton, R.S.; Barto, A.G. *Reinforcement Learning: An Introduction*; MIT Press: Cambridge, MA, USA, 1998.

2.  Thompson, W.R. On the likelihood that one unknown probability exceeds another in view of the evidence of two samples. *Biometrika* **1933**, *25*, 285–294.

3.  Robbins, H. Some aspects of the sequential design of experiments. *Bulletin of the American Mathematical Society* **1952**, *58*, 527–535.

4.  Bubeck, S.; Cesa-Bianchi, N. Regret analysis of stochastic and non-stochastic multi-armed bandit problems. *Foundations and Trends in Machine Learning* **2012**, *5*, 1–122.

5.  Lai, T.L.; Robbins, H. Asymptotically efficient adaptive allocation rules. *Advances in Applied Mathematics* **1985**, *6*, 4–22.

6.  Auer, P. Using confidence bounds for exploration-exploitation trade-offs. *Journal of Machine Learning Research* **2002**, *3*, 397–422.

7.  Auer, P.; Ortner, R. UCB revisited: Improved regret bounds for the stochastic multi-armed bandit problem. *Periodica Mathematica Hungarica* **2010**, *61*, 55–65.

8.  Garivier, A.; Cappé, O. The KL-UCB algorithm for bounded stochastic bandits and beyond. Proceedings of the Conference on Learning Theory (COLT); , 2011; pp. 359–376.

9.  Cappé, R.; Garivier, A.; Maillard, O.A.; Munos, R.; Stoltz, G. Kullback-Leibler upper confidence bounds for optimal sequential allocation. *Annals of Statistics* **2013**, *41*, 1516–1541.

10. Agarwal, S.; Goyal, N. Analysis of Thompson sampling for the multi-armed bandit problem. *Journal of Machine Learning Research* **2012**, *23*, 1–39.

11. Honda, J.; Takemura, A. An asymptotically optimal policy for finite support models in the multiarmed bandit problem. *Machine Learning* **2011**, *85*, 361–391.

12. Honda, J.; Takemura, A. Non-asymptotic analysis if a new bandit algorithm for semi-bounded rewards. *Journal of Machine Learning Research* **2015**, *16*, 3721–3756.

13. Auer, P.; Cesa-Bianchi, N.; Fischer, P. Finite-time analysis of the multi-armed bandit problem. *Machine Learning* **2002**, *47*, 235–256.

14. McMahan, H.B.; Streeter, M. Tight bounds for multi-armed bandits with expert advice. Proceedings of the Conference on Learning Theory (COLT); , 2009; pp. 1–10.

15. Stratonovich, R.L. On value of information. *Izvestiya of USSR Academy of Sciences, Technical Cybernetics* **1965**, *5*, 3–12.

16. Stratonovich, R.L.; Grishanin, B.A. Value of information when an estimated random variable is hidden. *Izvestiya of USSR Academy of Sciences, Technical Cybernetics* **1966**, *6*, 3–15.

17. Sledge, I.J.; Príncipe, J.C. Analysis of agent expertise in Ms. Pac-Man using value-of-information-based policies. *IEEE Transactions on Computational Intelligence and Artificial Intelligence in Games* **2017**. (under review).

18. Sledge, I.J.; Príncipe, J.C. Partitioning relational matrices of similarities or dissimilarities using the value of information. Proceedings of the IEEE International Conference on Acoustics, Speech, and Signal Processing (ICASSP); , 2018; pp. 1–5.

19. Sledge, I.J.; Emigh, M.S.; Príncipe, J.C. Guided policy exploration for Markov decision processes using an uncertainty-based value-of-information criterion. *IEEE Transactions on Neural Networks and Learning Systems* **2017**. (under review).

20. Cesa-Bianchi, N.; Fischer, P. Finite-time regret bounds for the multi-armed bandit problem. Proceedings of the International Conference on Machine Learning (ICML); , 1998; pp. 100–108.

21. Cesa-Bianchi, N.; Gentile, C.; Neu, G.; Lugosi, G. Boltzmann exploration done right. In *Advances in Neural Information Processing Systems (NIPS)*; Guyon, I.; Luxburg, U.V.; Bengio, S.; Wallach, H.; Fergus, R.; Vishwanathan, S.; Garnett, R., Eds.; MIT Press: Cambridge, MA, USA, 2017; pp. 6287–6296.

22. Kaelbling, L.P.; Littman, M.L.; Moore, A.W. Reinforcement learning: A survey. *Journal of Artificial Intelligence Research* **1996**, *4*, 237–285.

23. Salganicoff, M.; Ungar, L.H. Active exploration and learning in real-valued spaces using multi-armed bandit allocation indices. Proceedings of the International Conference on Machine Learning (ICML); , 1995; pp. 480–487.

24. Auer, P.; Cesa-Bianchi, N.; Freund, Y.; Schapire, R.E. The nonstochastic multi-armed bandit problem. *SIAM Journal on Computing* **2002**, *32*, 48–77.

25. Strehl, A.L.; Mesterharm, C.; Littman, M.L.; Hirsh, H. Experience-efficient learning in associative bandit problems. Proceedings of the International Conference on Machine Learning (ICML); , 2006; pp. 889–896.

26. Madani, O.; Lizotte, S.J.; Greiner, R. The budgeted multi-armed bandit problem. Proceedings of the Conference on Learning Theory (COLT); , 2004; pp. 643–645.

27. 　　Kleinberg, R.D. Nearly tight bounds for the continuum-armed bandit problem. In *Advances in Neural Information Processing Systems (NIPS)*; Saul, L.K.; Weiss, Y.; Bottou, L., Eds.; MIT Press: Cambridge, MA, USA, 2008; pp. 697–704.

28. 　　Wang, Y.; Audibert, J.; Munos, R. Algorithms for infinitely many-armed bandits. In *Advances in Neural Information Processing Systems (NIPS)*; Koller, D.; Schuurmans, D.; Bengio, Y.; Bottou, L., Eds.; MIT Press: Cambridge, MA, USA, 2008; pp. 1729–1736.

29. 　　Bubeck, S.; Munos, R.; Stoltz, G. Pure exploration in finitely-armed and continuous-armed bandits. *Theoretical Computer Science* **2011**, *412*, 1876–1902.

30. 　　Vermorel, J.; Mohri, M. Multi-armed bandit algorithms and empirical evaluation. In *Machine Learning: ECML*; Gama, J.; Camacho, R.; Brazdil, P.B.; Jorge, A.M.; Torgo, L., Eds.; Springer-Verlag: New York City, NY USA, 2005; pp. 437–448.

31. 　　Even-Dar, E.; Mannor, S.; Mansour, Y. PAC bounds for multi-armed bandit and Markov decision processes. Proceedings of the Conference on Learning Theory (COLT); , 2002; pp. 255–270.

32. 　　Mannor, S.; Tsitsiklis, J.N. The sample complexity of exploration in the multi-armed bandit problem. *Journal of Machine Learning Research* **2004**, *5*, 623–648.

33. 　　Langford, J.; Zhang, T. The epoch-greedy algorithm for multi-armed bandits. In *Advances in Neural Information Processing Systems (NIPS)*; Platt, J.C.; Koller, D.; Singer, Y.; Roweis, S.T., Eds.; MIT Press: Cambridge, MA, USA, 2008; pp. 817–824.

34. 　　Srinivas, S.; Krause, A.; Seeger, M.; Kakade, S.M. Gaussian process optimization in the bandit setting: No regret and experimental design. Proceedings of the International Conference on Machine Learning (ICML); , 2010; pp. 1015–1022.

35. 　　Krause, A.; Ong, C.S. Contextual Gaussian process bandit optimization. In *Advances in Neural Information Processing Systems (NIPS)*; Shawe-Taylor, J.; Zemel, R.S.; Bartlett, P.L.; Pereira, F.; Weinberger, K.Q., Eds.; MIT Press: Cambridge, MA, USA, 2011; pp. 2447–2455.

36. 　　Beygelzimer, A.; Langford, J.; Li, L.; Reyzin, L.; Schapire, R.E. Contextual bandit algorithms with supervised learning guarantees. Proceedings of the International Conference on Artificial Intelligence and Statistics (AISTATS); , 2011; pp. 19–26.

37. 　　Sledge, I.J.; Príncipe, J.C. Balancing exploration and exploitation in reinforcement learning using a value of information criterion. Proceedings of the IEEE International Conference on Acoustics, Speech, and Signal Processing (ICASSP); , 2017; pp. 1–5.

38. 　　Thathachar, K.S.N.M.A.L. Learning automata–A survey. *IEEE Transactions on Systems, Man, and Cybernetics* **1974**, *4*, 323–334.

39. 　　Thathachar, M.A.L.; Sastry, P.S. A new approach to the design of reinforcement schemes for learning automata. *IEEE Transactions on Systems, Man, and Cybernetics* **1985**, *15*, 168–175.

40. 　　Audibert, J.Y.; Munos, R.; Szepesvári, C. Exploration-exploitation tradeoff using variance estimates in multi-armed bandits. *Theoretical Computer Science* **2009**, *410*, 1876–1902.

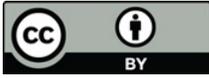

Table 1: Variable Notation

| Symbol | Description |
|---|---|
| $\mathcal{R}$ | Continuous space of real values |
| $\mathcal{A}$ | Discrete space of slot-machine arms |
| $|\mathcal{A}|$ | Total number of slot-machine arms |
| $T$ | Total number of arm pulls |
| $s, k$ | A given arm pull index in the range $[0, T]$; index to a particular iteration or round |
| $a^i$ | A given slot-machine arm, indexed by $i$; $i$ takes values in the range $[1, |\mathcal{A}|]$ |
| $a^*$ | The slot-machine arm that yields the best expected reward |
| $a_k^i$ | A particular slot machine arm $i$ pulled at round $k$ |
| $X_k^i$ | The real-valued reward obtained by pulling arm $a^i$ at round $k$; can also be interpreted as a random variable |
| $\mu^i$ | The expected reward for a non-optimal slot machine arm $a^i$ |
| $\mu^*$ | The expected reward for the optimal slot machine arm $a^*$ |
| $\sigma^2$ | The variance of all slot machine arms |
| $p_0(a_k^i)$ | A prior probability distribution that specifies the chance of choosing arm $a_k^i$ at round $k$ |
| $p(a_k^i)$ | A posterior-like probability distribution that specifies the chance of choosing arm $a^i$ at round $k$ |
| $p(a^j)$ | A probability distribution that specifies the chance of choosing arm $a^j$ independent of a given round $k$ |
| $\pi_k^i$ | Equivalent to $p(a_k^i)$; represents the arm-selection policy |
| $\mathbb{E}.[\cdot]$ | The expected value of random variable according to a probability distribution, e.g., $\mathbb{E}_{a_k^i}[X_k^i]$ represents $\sum_{a_k^i \in \mathcal{A}} p(a_k^i) X_k^i$ |
| $D_{\mathrm{KL}}(\cdot \| \cdot)$ | The Kullback-Leibler divergence between two probability distributions, e.g., $D_{\mathrm{KL}}(p(a_k^i) \| p_0(a_k^i))$ is the divergence between $p(a_k^i)$ and $p_0(a_k^i)$ |
| $\varphi_{\inf}$ | A positive, real-valued information constraint bound |
| $\varphi_{\mathrm{cost}}$ | A positive, real-valued reward constraint bound |

| Symbol | Description |
|---|---|
| $q_{k+1}^i$ | An estimate of the expected reward for slot machine arm $a^i$ at round $k+1$ |
| $I_k^i$ | A binary-valued indicator variable that specifies if slot machine arm $a^i$ was chosen at round $k$ |
| $\epsilon$ | A non-negative, real-valued parameter that dictates the amount of exploration for the $\epsilon$-greedy algorithm (GreedyMix) |
| $\tau$ | A non-negative, real-valued parameter that dictates the amount of exploration for soft-max-based selection (SoftMix) and value-of-information-based exploration (VoIMix) |
| $\gamma$ | A non-negative, real-valued parameter that represents the mixing coefficient for value-of-information-based exploration (VoIMix and AutoVoIMix) |
| $\theta$ | A non-negative, real-valued hyperparameter that dictates the exploration duration for value-of-information-based exploration (AutoVoIMix) |
| $d$ | A positive, real-valued hyperparameter that dictates the exploration duration for value-of-information-based exploration (VoIMix) |
| $\tau_k, \gamma_k, \theta_k, d_k$ | The values of $\tau$, $\gamma$, $\theta$, and $d$, respectively, at a given round $k$ |
| $\delta_i$ | The mean loss incurred by choosing arm $a^i$ instead of arm $a^*$: $\delta_i = \mu^* - \mu^i$ |
| $V_p^j$ | A random variable representing the ratio of the reward for a given slot machine arm to its chance of being selected: $X_p^j / \pi_p^j$ |
| $Z_s^i$ | A random variable representing the expression $\delta_i + V_s^i - V_s^*$ |
| $Q_{k-1}^i$ | A random variable representing the expression $\prod_{s=1}^{k-1} \exp(\tau_s V_s^i - \tau_s V_s^*)$ |
| $\mathcal{J}_s$ | A $\sigma$-algebra: $\sigma(X_s^i, 1 \le s \le k), k \ge 1$ |
| $\phi_c(\tau_s)$ | A real-valued function parameterized by $c$: $\phi_c(\tau_s) = (e^{c\tau_s} - c\tau_s - 1)/\tau_s^2$ |
| $\omega, \omega'$ | Real values inside the unit interval |
| $c_k$ | A real-valued sequence $1 + 2|\mathcal{A}|/\gamma_k$ for VoIMix or $1 + |\mathcal{A}|/\gamma_k$ for AutoVoIMix |
| $\sigma_k^2$ | A real-valued sequence $-d^2 + 2|\mathcal{A}|/\gamma_k$ for VoIMix or $2|\mathcal{A}|/\gamma_k$ for AutoVoIMix |